\documentclass[runningheads]{llncs}

\usepackage{amsfonts, amsmath, amsthm}
\usepackage[T1]{fontenc}
\usepackage{microtype}
\usepackage{graphicx}
\usepackage{subfigure}
\usepackage{booktabs} 
\usepackage{makecell}
\usepackage{graphicx}
\usepackage[misc]{ifsym}
\usepackage[lined,linesnumbered,ruled,commentsnumbered,resetcount,vlined]{algorithm2e}

\newcommand\SVDeq{\mathrel{\stackrel{\makebox[0pt]{\mbox{\normalfont\tiny SVD}}}{\,=\,}}}
\newcommand\FULLSVDeq{\mathrel{\stackrel{\makebox[0pt]{\mbox{\normalfont\tiny FULL-SVD}}}{\,\,\,\,\,\,=\,\,\,\,\,\,}}}
\newcommand\RSVDapprox{\mathrel{\stackrel{\makebox[0pt]{\mbox{\normalfont\tiny RSVD}}}{\,\,\approx\,\,}}}
\newcommand{\norm}[1]{\left\lVert#1\right\rVert}

\begin{document}
	
\newtheorem*{Proposition_3_1}{Proposition 3.1: Bounds describing eigenvalue decay of $\bar {\mathcal M_k}$}
%
\title{Randomized K-FACs: Speeding up K-FAC with Randomized Numerical Linear Algebra}
\titlerunning{Randomized K-FACs: Speeding up K-FAC with r-NLA}
%
%
\author{Constantin Octavian Puiu\orcidID{0000-0002-1724-4533} \Letter}

%
\institute{University of Oxford, Mathematical Institute,\\
	\email{constantin.puiu@maths.ox.ac.uk}\\
}

\toctitle{Randomized K-FACs: Speeding up K-FAC with Randomized Numerical Linear Algebra}
\tocauthor{Constantin~Octavian~Puiu}

\authorrunning{C. O. Puiu}

\maketitle   

%



\begin{abstract}
	\textsc{k-fac} is a successful tractable implementation of Natural Gradient for Deep Learning, which nevertheless suffers from the requirement to compute the inverse of the Kronecker factors (through an eigen-decomposition). This can be very time-consuming (or even prohibitive) when these factors are large. In this paper, we theoretically show that, owing to the exponential-average construction paradigm of the Kronecker factors that is typically used, their eigen-spectrum must decay. We show numerically that in practice this decay is very rapid, leading to the idea that we could save substantial computation by only focusing on the first few eigen-modes when inverting the Kronecker-factors. Importantly, the spectrum decay happens over a constant number of modes irrespectively of the layer width. This allows us to reduce the time complexity of \textsc{k-fac} from cubic to quadratic in layer width, partially closing the gap w.r.t.\ \textsc{seng} (another practical Natural Gradient implementation for Deep learning which scales linearly in width). Randomized Numerical Linear Algebra provides us with the necessary tools to do so. Numerical results show we obtain $\approx2.5\times$ reduction in per-epoch time and $\approx3.3\times$ reduction in time to target accuracy. We compare our proposed \textsc{k-fac} sped-up versions \textsc{seng}, and observe that for \textit{CIFAR10} classification with \textit{VGG16\_bn} we perform on par with it.
\end{abstract}
\section{Introduction}
Research in optimization for DL has lately focused on Natural Gradient (NG), owing to its desirable properties when compared to standard gradient \cite{AMARI,New_insights_and_perspectives}. \textsc{k-fac} (\cite{KFAC}) is a \textit{tractable} implementation which nevertheless suffers from the drawback of requiring the actual inverses of the Kronecker Factors (not just a linear solve). This computation scales \textit{cubically} in layer width. When K-Factors are large (eg.\ for very wide fully-connected layers), \textsc{k-fac} becomes very slow. A fundamentally different practical implementation of NG without this problem has been proposed: \textsc{seng} \cite{SENG} (uses matrix sketching  \cite{Sketch_main_paper} and empirical NG \cite{New_insights_and_perspectives}). \textsc{seng} scales linearly in layer width, thus substantially outperforming \textsc{k-fac} for very wide nets.

In this paper, we provide a way to alleviate \textsc{k-fac}'s issue and make it competitive with \textsc{seng}, by partly closing the \textit{complexity gap}. We begin by theoretically noting that the eigenspectrum of the K-Factors must decay rapidly, owing to their exponential-average (EA) construction paradigm. Numerical results of practically obtained eigen-spectrums show that in practice, the decay is much faster than the one implied by our worst-case scenario theoretical analysis. Using these observations, we employ randomized Numerical Linear Algebra (rNLA, \cite{RSVD_paper}) to reduce the time complexity \textit{from cubic to quadratic} in layer width. This gives us highly time-efficient approximation routes for \text{K-Factors} inversion, with minimal accuracy reduction. Numerically, our proposed methods speed up \textsc{k-fac} by $2.5\times$ and $3.3\times$ in terms of \textit{time per epoch} and \textit{time to target accuracy} respectively. Our algorithms outperform \textsc{seng} \cite{SENG} (w.r.t.\ wall time) for moderate and high target test accuracy, but slightly underperform for very high test accuracy.

\textbf{Related Work:}
The work of Tang et.\ al.\ (2021, \cite{low_rank_Kfac_SKFAC}) is most related. However, their main approach is to construct a more efficient inversion of the regularized low-rank K-factors, without any rNLA. To make their approach feasible, they have to perform an EA over $A_k^{(l)}$ and $G_k^{(l)}$ rather than over $\bar {\mathcal A}_k^{(l)}$ and $\bar \Gamma_k^{(l)}$, as is standard (see \textit{Section 2.1}). Our approach avoids this issue. Osawa et.\ al.\ (2020, \cite{Kazuki_SPNGD}) presents some ideas to speed-up \textsc{k-fac}, but they are orthogonal to ours.

\section{Preliminaries}
\textbf{Neural Networks (NNs): }
Our learning problem is 
\begin{equation}
\min_{\theta} f(\theta) := \frac{1}{|\mathcal D|}\sum_{(x_i,y_i)\in\mathcal D}\big(- \log p (y_i|h_\theta(x_i))\big),
\label{eqn_optimization_problem}
\end{equation}

\noindent where $\mathcal D$ is the dataset containing input-target pairs $\{x_i,y_i\}$, $\theta$ are the aggregated network parameters, $h_{\theta}(\cdot)$ is the neural network function (with $n_L$ layers), and $p(y|h_\theta(x_i))$ is the predictive distribution of the network (over labels - e.g.\ over classes), which is parameterized by $h_\theta(x_i)$. We let $p_\theta (y|x):=p (y|h_\theta(x))$, $g_k:=\nabla_\theta f(\theta_k)$, and note that we can express $g_k = [g_k^{(1)}, ..., g_k^{(n_L)}]$, where $g_k^{(l)}$ is the gradient of parameters in layer $l$. We will always use a superscript to refer to the \textit{layer} index and a subscript to refer to the \textit{optimization iteration} index. 

\subsection{Fisher Information, Natural Gradient and K-FAC}
The Fisher information is defined as
\begin{equation}
F_k := F(\theta_k) := \mathbb E_{\substack{ x\sim \mathcal D \\ y\sim p_\theta(y|x)}}\biggl[\nabla_\theta \log p_\theta(y|x) \nabla_\theta \log p_\theta(y|x)^T\biggr].
\label{eqn_FIM}
\end{equation}
A NG descent (NGD) algorithm with stepsize $\alpha_k$ takes steps of the form $s^{(\text{NGD})}_k = - \alpha_k \nabla_{\text{NG}} f(\theta_k)$, where $\nabla_{\text{NG}} f(\theta_k)$ is the natural gradient (NG), defined as \cite{AMARI}
\begin{equation}
\nabla_{NG} f(\theta_k) := F_k^{-1}g_k.
\end{equation}
In DL, the dimension of $F_k$ is very large, and $F_k$ can neither be stored nor used to complete a linear-solve. \textsc{k-fac} (\cite{KFAC}) is a practical implementation of the NGD algorithm which bypasses this problem by approximating $F_k$ as
\begin{equation}
F_k^{(\text{KFAC})} := \text{blockdiag}\big(\{\mathcal A^{(l)}_k \otimes \Gamma^{(l)}_k \}_{l=1,...,n_L}\big),
\label{KFAC_matrix_defn}
\end{equation}
where $\mathcal A^{(l)}_k:= A_k^{(l)}[A_k^{(l)}]^T$ and $\Gamma^{(l)}_k:= G_k^{(l)}[G_k^{(l)}]^T$ are the \textit{forward K-factor} and \textit{backward K-factor} respectively (of layer $l$ at iteration $k$) \cite{KFAC}. Each block corresponds to a layer and $\otimes$ denotes the Kronecker product. The exact K-Factors definition depends on the layer type (see \cite{KFAC} for FC layers, \cite{convolutional_KFAC} for Conv layers). For our purpose, it is sufficient to state that $A_k^{(l)}\in\mathbb R^{d^{(l)}_{\mathcal A} \times n^{(l)}_A}$ and $G_k^{(l)}\in\mathbb R^{d^{(l)}_{\Gamma}\times n^{(l)}_\Gamma}$, with $n^{(l)}_{\mathcal A}, n^{(l)}_\Gamma \propto n_{\text{BS}}$, where $n_{\text{BS}}$ is the batch size (further size details in \cite{KFAC,convolutional_KFAC}).

Computing $(F_k^{(\text{KFAC})})^{-1}g_k$ can be done relatively efficiently in a block-wise fashion, since we have $(\mathcal A^{(l)}_k \otimes \Gamma^{(l)}_k)^{-1}g_k^{(l)} = \text{vec}\big([\Gamma^{(l)}_k]^{-1} \text{Mat}(g_k^{(l)}) [\mathcal A^{(l)}_k]^{-1}\big)$, where $\text{vec}(\cdot)$ is the matrix vectorization operation and $\text{Mat}(\cdot)$ is its inverse. Note that since $\text{Mat}(g_k^{(l)})$ is a matrix, we need to \textit{compute} the inverses of $\bar{\mathcal A}^{(l)}_k$ and $\bar \Gamma^{(l)}_k$ (eg.\ through an eigen-decomposition - and not just linear-solve with them). This is point is essential.

\textsc{k-fac} pseudo-code is shown in \textit{Algorithm 1}. Note that in practice, instead of assembling $F_k^{(\text{KFAC})}$ as in equation (\ref{KFAC_matrix_defn}), with the K-factors local to $\theta_k$ ($\mathcal A^{(l)}_k$ and $\Gamma^{(l)}_k$), we use an exponential average (EA) ($\bar{\mathcal A}^{(l)}_k$ and $\bar \Gamma^{(l)}_k$; see \textit{lines 4} and \textit{8} in \textit{Algorithm 1}). This aspect is important for our discussion in \textit{Section 3}. In \textit{Algorithm 1} we initialize $\bar {\mathcal A}^{(l)}_{-1} := I$ and $\bar \Gamma^{(l)}_{-1} := I$. $\theta_0$ is initialized as typical \cite{Initialization}.

\begin{algorithm}[H]
	\footnotesize
	\caption{\textsc{k-fac} \cite{KFAC}}
	
	\For{$k=0,1,2,....$, $\text{with sampled batch}$ $\mathcal B_k \subset \mathcal D$}{
		
		\For(\tcp*[h]{Perform forward pass}){$l=0,1,...,N_L$}
		{
			Get $a^{(l)}_k$ and $A^{(l)}_k$
			
			$\bar {\mathcal A}^{(l)}_k \leftarrow \rho \bar {\mathcal A}^{(l)}_{k-1} + (1-\rho) A^{(l)}_k [A^{(l)}_k]^T$ \tcp{Update fwd.\ EA K-factors}
			
		}
		Get $\tilde f(\theta_k)$ \tcp*{The batch-estimate of $f(\theta_k)$, from $a^{(l)}_k$}
		
		\For(\tcp*[h]{Perform backward pass}){$l=N_L,N_{L-1},...,1$}
		{
			Get $g^{(l)}_k$ and $G^{(l)}_k$
			
			$\bar \Gamma^{(l)}_k \leftarrow \rho \bar \Gamma^{(l)}_{k-1} + (1-\rho) G^{(l)}_k [G^{(l)}_k]^T$ \tcp{Update bwd.\ EA K-factors}
		}
		Get gradient $g_k = \big[\big(g^{(1)}_k\big)^T,... \big(g^{(N_L)}_k\big)^T\big]^T$
		
		\For(\tcp*[h]{Compute \textsc{k-fac} step:}){$l=0,1,...,N_L$}
		{
			\tcp{Get Eig of $\bar {\mathcal A}^{(l)}_k$ and $\bar \Gamma^{(l)}_k$ for inverse application}
			$U^{(l)}_{A,k} D^{(l)}_{A,k} \big(U^{(l)}_{A,k}\big)^T = \text{eig}(\bar {\mathcal A}^{(l)}_k)$ ; $U^{(l)}_{\Gamma,k} D^{(l)}_{\Gamma,k} \big(U^{(l)}_{\Gamma,k}\big)^T = \text{eig}(\bar \Gamma^{(l)}_k)$
			
			\tcp{Use Eigs to apply K-FAC EA matrices inverses to $g^{(l)}_k$}
			
			$M^{(l)}_k  = \text{Mat}(g^{(l)}_k)  U^{(l)}_{A,k} (D^{(l)}_{A,k}+ \lambda I)^{-1} \big(U^{(l)}_{A,k}\big)^T $
			
			$S^{(l)}_k = U^{(l)}_{\Gamma,k} (D^{(l)}_{\Gamma,k}+ \lambda I)^{-1} \big(U^{(l)}_{\Gamma,k}\big)^T M^{(l)}_k  $ ; $s^{(l)}_k = \text{vec}(S^{(l)}_k)$

		}
		$\theta_{k+1} = \theta_k - \alpha_k[(s_k^{(1)})^T,...,(s_k^{(N_L)})^T]^T$ 	\tcp{Take K-FAC step}
		
	}	

\end{algorithm}

\subsubsection{Key notes on Practical Considerations}
In practice, we update the Kronecker-factors and recompute their eigendecompositions (``inverses'') only every few tens/hundreds of steps (update period $T_{K,U}$, inverse computation period $T_{K,I}$) \cite{KFAC}. Typically, we have $T_{K,I}> T_{K,U}$. As we began in \textit{Algorithm 1}, we formulate our discussion for the case when $T_{K,I}= T_{K,U}=1$. We do this purely for simplicity of exposition\footnote{To avoid \textit{if} statements in the presented algorithm.}. Extending our simpler discussion to the case when these operations happen at a smaller frequency is \textit{trivial}, and does not modify our conclusions. Our practical implementations use the standard practical procedures.


\subsection{Randomized SVD (RSVD)}
Before we begin diving into rNLA, we note that whenever we say \textsc{rsvd}, or \textsc{qr}, we always refer to the thin versions unless otherwise specified. Let us focus on the arbitrary matrix $X\in\mathbb R^{m\times n}$. For convenience, assume for this section that $m>n$ (else we can transpose $X$). Consider the \textsc{svd} of $X$
\begin{equation}
X \SVDeq U_X \Sigma_X V_X^T ,
\label{eqn_SVD_of_X}
\end{equation}
and assume $\Sigma_X$ is sorted decreasingly. It is a well-known fact that the best\footnote{As defined by closeness in the ``$(p,k)$-norm'' (see for example \cite{Optimality_of_SVD}).} rank-$r$ approximation of $X$ is given by $U_X[:,:r] \Sigma_X[:r,:r]V_X[:,:r]^T$ \cite{RSVD_paper}. The idea behind randomized \textsc{svd} is to obtain these first $r$ singular modes without computing the entire (thin) \textsc{svd} of $X$, which is $\mathcal O(mn^2)$ time complexity. \textit{Algorithm 2} shows the \textsc{rsvd} algorithm alongside with associated time complexities. We omit the derivation and error analysis for brevity (see \cite{RSVD_paper} for details).

\begin{algorithm}[H]
	\footnotesize
	\caption{Randomized \textsc{svd} (\textsc{rsvd}) \cite{RSVD_paper}}
	\textbf{Input:} $X \in \mathbb R^{m\times n}$, target rank $r < \min(m,n)$, oversampling param.\ $r_l \leq n-r$
	
	\textbf{Output:} Approximation of the first $r$ singular modes of $X$
	
	Sample Gaussian Matrix $\Omega \in \mathbb R^{n\times (r+l)}$ \tcp{$\mathcal O(n(r+l))$ flops}
	
	Compute $X\Omega$ \tcp{$\mathcal O(mn(r+l))$ flops}
	
	$Q R = \text{QR\_decomp} (X\Omega)$ \tcp{$\mathcal O (m(r+l)^2)$ flops}
	
	$B := Q^TX\in\mathbb R^{(r+l)\times n}$ \tcp{$\mathcal O(nm(r+l))$ flops}
	
	Compute Full SVD (i.e. not the thin one) of $B^T$, and transpose it to recover $B \,\,\,\FULLSVDeq\,\,\, U_B \Sigma_B V_B^T$ \tcp{$\mathcal O(n^2(r+l))$ flops}
	
	$\tilde U_X = QU_B$; $\tilde \Sigma_X = \Sigma_B[:r,:r]$; $\tilde V_X = V_B[:,:r]$ \tcp{$\mathcal O(m(r+l)^2)$ flops}
	
	\textbf{Return} $\tilde U_X\in \mathbb R^{m\times r}$, $\tilde \Sigma_X\in \mathbb R^{r\times r}$, $\tilde V_X\in \mathbb R^{n\times r}$
\end{algorithm}

The returned quantities, $\tilde U_X\in \mathbb R^{m\times r}$, $\tilde \Sigma_X\in \mathbb R^{r\times r}$, $\tilde V_X\in \mathbb R^{n\times r}$ are approximations for $U_X[:,:r]$, $\Sigma_X[:r,:r]$ and $V_X[:,:r]$ respectively - which is what we were after. These approximations are relatively good with high probability, particularly when the singular values spectrum is rapidly decaying \cite{RSVD_paper}. The total complexity of \textsc{rsvd} is $\mathcal O(n^2(r+r_l) + mn(r+r_l))$ - significantly better than the complexity of \textsc{svd} $\mathcal O(m^2n)$ when $r+r_l\ll \min(m,n)$. We will see how we can use this to speed up \textsc{k-fac} in \textit{Section 4.1}. Note the presence of the over-sampling parameter $r_l$, which helps with accuracy at minimal cost. This $r_l$ will appear in many places. Finally, we note that $Q$ is meant to be a skinny-tall orthonormal matrix s.t.\ $\norm{X-QQ^TX}_F$ is ``small''. There are many ways to obtain $Q$, but in \textit{lines 3-4} of \textit{Algorithm 2} we presented the simplest one for brevity (see \cite{RSVD_paper} for details). In practice we perform the power iteration in \textit{line 4} $n_{\text{pwr-it}}$ times (possibly more than once).

\subsubsection{RSVD Error Components}
Note that there are two error components when using the returned quantities of an \textsc{rsvd} to approximate a matrix. The first component is the \textit{truncation error} - which is the error we would have if we computed the \textsc{svd} and then truncated. The second error is what we will call \textit{projection error}, which is the error between the rank-$r$ \textsc{svd}-truncated $X$ and the \textsc{rsvd} reconstruction of $X$ (which appears due to the random Gaussian matrix). 

\subsubsection{RSVD for Square Symmetric PSD matrices}
When our matrix $X$ is square-symmetric PSD (the case we will fall into) we have $m=n$, $U_X = V_X$, and the \textsc{svd} (\ref{eqn_SVD_of_X}) is also the eigen-value decomposition. As \textsc{rsvd} brings in significant errors\footnote{Relatively small, but higher than machine precision - as SVD would have.}, \textit{Algorithm 2} will return $\tilde U_X\ne \tilde V_X$ even in this case. Thus, we have to choose between using $\tilde V_X$ and $\tilde U_X$ (or any combination of these). A key point to note is that $\tilde V_X$ approximates $V_X[:,:r]$ better than $\tilde U_X$ approximates $U_X[:,:r]$ \cite{RSVD_theory_error}. Thus using $\tilde V_X\tilde \Sigma_X\tilde V_X^T$  as the rank-$r$ approximation to $X$ is preferable. This is what we do in practice, and it gives us virtually zero \textit{projection error}.

\subsection{Symmetric Randomized EVD (SREVD)}
When $X$ is square-symmetric PSD we have $m=n$, $U_X = V_X$, and the \textsc{svd} (\ref{eqn_SVD_of_X}) is also the eigen-value decomposition (\textsc{evd}). In that case, we can exploit the symmetry to reduce the computation cost of obtaining the first $r$ modes. \textsc{srevd} is shown in \textit{Algorithm 3}. The returned quantities, $\tilde U_X\in \mathbb R^{m\times r}$ and $\tilde \Sigma_X\in \mathbb R^{r\times r}$ are approximations for $U_X[:,:r]$ and $\Sigma_X[:r,:r]$ respectively - which is what we were after. The same observations about $Q$ that we made in \textit{Section 2.2} also apply here.
\begin{algorithm}[H]
	\footnotesize
	\label{Symmetric_randomized_EVD}
	\caption{Symmetric Randomized \textsc{evd} (\textsc{srevd}) \cite{RSVD_paper}}
	\textbf{Input:} Square, Symmetric PSD matrix $X \in\mathbb R^{n \times n}$
	
	\textbf{Output:} Approximation of the first $r$ eigen-modes of $X$
	
	Sample Gaussian Matrix $\Omega \in \mathbb R^{n\times (r+l)}$ \tcp{$\mathcal O(n(r+l))$ flops}
	
	Compute $X\Omega$ \tcp{$\mathcal O(n(r+l))$ flops}
	
	$Q R = \text{QR\_decomp} (X\Omega)$ \tcp{$\mathcal O (n(r+l)^2)$ flops}
	
	Compute $C = Q^TXQ$ \tcp{$\mathcal O(n^2(r+l))$ flops}
	
	$P_C D_C P_C^T=\text{Eigen\_decomp}(C)$ \tcp{$\mathcal O((r+l)^3)$ flops}
	
	$\tilde U_X = QP_C$; $\tilde \Sigma_X \in\mathbb R^{r\times r}$ \tcp{$\mathcal O(n(r+l)^2)$ flops}
	
	\textbf{Return}   $\tilde U_X \in\mathbb R^{n\times r}$, $\tilde \Sigma_X \in\mathbb R^{r\times r}$ 
\end{algorithm}

The complexity is still $\mathcal O(n^2(r+l))$ as with \textsc{rsvd}\footnote{Set $m=d$ in \textsc{rsvd} complexity}, but the full-SVD of $(Q^TX)^T$ ($\mathcal O(n^2(r+l))$ complexity) is now replaced by a matrix-matrix multiplication of $\mathcal O(n(r+l)^2)$ and a virtually free eigenvalue decomposition. However, note that by projecting both the columnspace and the rowspace of $X$ onto $Q$, we are losing accuracy because we are essentially not able to obtain the more accurate $\tilde V_X$ as we did with \textsc{rsvd}. That is because we have $P_C = Q^TU_X$, and thus we can only obtain $\tilde U_X = QQ^TU_X$ but \textit{not} $\tilde V_X$. Consequently, the \textit{projection error} is larger for \textsc{srevd} than for \textsc{rsvd}, although the turncation error is the same. 

\section{The Decaying Eigen-spectrum of K-Factors}
\subsubsection{Theoretical Investigation} Let  $\lambda_M$ be the max.\ eigenvalue of the arbitrary EA Kronecker-factor
\begin{equation}
\bar {\mathcal M_k} = (1-\rho)\sum_{i=-\infty}^{k}\rho^{k-i}M_iM_i^T,
\label{matrix_form_EA}
\end{equation}

\noindent with $M_i\in\mathbb R^{d_M\times n_M}$, $n_M \propto n_{\text{BS}}$.  We now look at an upperbound on the number of eigenvalues that satisfy $\lambda_i\geq \epsilon \lambda_M$ (for some assumed $\alpha\in(0,1)$, chosen $\epsilon\in(0,1)$, and \textit{``sufficiently large''} given $d_M$). Proposition 3.1 gives the result.

\begin{Proposition_3_1}
	Consider the $\bar{\mathcal M}_k$ in (\ref{matrix_form_EA}), let $\lambda_M$ be its maximum eigenvalue, and let us choose some $\epsilon\in(0,1)$. Assume the maximum singular value of $M_i$ is $\leq\sigma_M$ $\forall k$, and that we have $\lambda_M \geq \alpha \sigma_M^2$ for some fixed $\alpha\in(0,1)$. Then, we have that at most $\min(r_\epsilon n_{M}, d_M)$ eigenvalues of $\bar{\mathcal M_k}$ are above $\epsilon \lambda_M$, with
	\begin{equation}
	r_\epsilon = \left \lceil{{\log(\alpha\epsilon)}/{\log(\rho)}}\right \rceil.
	\label{eqn_prop_3_1}
	\end{equation}
\end{Proposition_3_1}

\textit{Proof.} We have $\bar {\mathcal M_k} = \bar{\mathcal M}_{\text{old}} + \bar{\mathcal M}_{\text{new}}$ with $\bar{\mathcal M}_{\text{old}} := (1-\rho)\sum_{i=-\infty}^{k-r}\rho^{k-i}M_iM_i^T$, and $\bar{\mathcal M}_{\text{new}} := (1-\rho)\sum_{i=k-r+1}^{k}\rho^{k-i}M_iM_i^T$. 

First, let us find $r$ s.t.\ the following desired upper-bound holds:
\begin{equation}
\lambda_{\text{Max}}(\bar{\mathcal M}_{\text{old}}) \leq \alpha\epsilon \sigma_M^2.
\label{eqn_desired_eval_bound_prop_3_1}
\end{equation}
Let $\rho_C:=(1-\rho)$. By using $\norm{\cdot}_2=\lambda_{\text{Max}}(\cdot)$ for s.p.s.d.\ arguments, we have
\begin{equation}
\lambda_{\text{Max}}\big(\bar{\mathcal M}_{\text{old}}\big)\leq \rho_C\sum_{i=-\infty}^{k-r}\rho^{k-i}\lambda_{\text{Max}}\big(M_iM_i^T\big) \leq \rho_C \sigma_M^2\rho^r\sum_{i=0}^\infty \rho^i = \sigma_M^2 \rho^r.
\label{upperbound_M_bar_old}
\end{equation}

Thus, in order to get (\ref{eqn_desired_eval_bound_prop_3_1}) to hold, we can set $\sigma_M^2\rho^r\leq \epsilon\alpha \sigma_M^2$ from (\ref{upperbound_M_bar_old}). That is, we must have $r\geq\log(\alpha\epsilon)/\log(\rho)$. Thus, choosing 
\begin{equation}
r:=r_{\epsilon}:=\left \lceil{{\log(\alpha\epsilon)}/{\log(\rho)}}\right \rceil
\end{equation}
ensures (\ref{eqn_desired_eval_bound_prop_3_1}) holds. Now, clearly, $\text{rank}\big(\bar{\mathcal M}_{\text{new}}\big) \leq n_{M}r$, so $\bar{\mathcal M}_{\text{new}}$ has at most $n_{M}r$ non-zero eigenvalues. Using $\bar {\mathcal M_k} = \bar{\mathcal M}_{\text{old}} + \bar{\mathcal M}_{\text{new}}$ and the upperbound (\ref{eqn_desired_eval_bound_prop_3_1}) (which holds for our choice of $r=r_\epsilon$) gives that $\bar {\mathcal M_k}$ has at most $n_{M}r_\epsilon$ eigenvalues above $\alpha\epsilon \sigma_M^2$. But by assumption the biggest eigenvalue of $\bar {\mathcal M_k}$ satisfies $\lambda_M\geq \alpha \sigma_M^2$. Thus, at most $n_{M}r_\epsilon$ of $\bar {\mathcal M_k}$ satisfy $\lambda_i\geq \epsilon \lambda_M$. This completes the proof. $\square$

\textit{Proposition 3.1} gives the notable result\footnote{Although from the perspective of a fairly loose bound.} that the number of modes we need to save for a target $\epsilon$ depends only on our tolerance level $\epsilon$ (practically $\epsilon=1/33$) and on the batch-size (through $n_M\propto n_{\text{BS}}$), but \textit{not} on $d_M$. To see this, note that \textit{Proposition 3.1} gives that the number of modes to save is in principle\footnote{Assuming it does not exceed $d_M$ in which case it becomes $d_M$.} $r_{\epsilon}n_M = \left \lceil{{\log(\alpha\epsilon)}/{\log(\rho)}}\right \rceil n_{M}$, which does not depend on $d_M$. Thus, increasing $d_M$ (past $r_{\epsilon}n_M$) does not affect how many modes we need to compute (to ensure we only ignore eigenvalues below $\epsilon\lambda_{\text{max}}$). Intuitively, this means we can construct approaches which scale better in $d_M$ than \textsc{evd}: the \textsc{evd} computes $d_M$ modes when we only really need a constant (w.r.t.\ $d_M$) number of modes\footnote{Thus, for large $d_M$ most of the computed eigen-modes are a waste!}! This is good news for \textsc{k-fac}: its bottleneck was the scaling of \textsc{evd} with the net width ($d_M$'s)!

The assumption about $\lambda_M$ may seem artificial, but holds well in practice. A more in depth analysis may avoid it. Plugging realistic values of $\epsilon = 0.03$, $\alpha = 0.1$ and $\rho=0.95$, $n_M = n_{\text{BS}}=256$ (holds for FC layers) in \textit{Proposition 3.1} tells us we have to retain at least $n_{M} r_{\epsilon}=29184$ eigenmodes to ensure we only ignore eigenvalues satisfying $\lambda_i \leq 10^{-1.5}\lambda_M$. Clearly, 29184 is very large, and \textit{Proposition 3.1} is not directly useful in practice. However, it does ensure us that the eigenspectrum of the EA K-Factors must have a form of which implies we only really need to keep a constant number (w.r.t.\ $d_M$) of eigenmodes. We now show numerically that this decay is much more rapid than inferred by our worst-case analysis here.

\begin{figure}[t]
	\centering
	
	\includegraphics[trim={0.1cm 0.1cm 1.5cm 0.4cm},clip,width=0.323\textwidth]{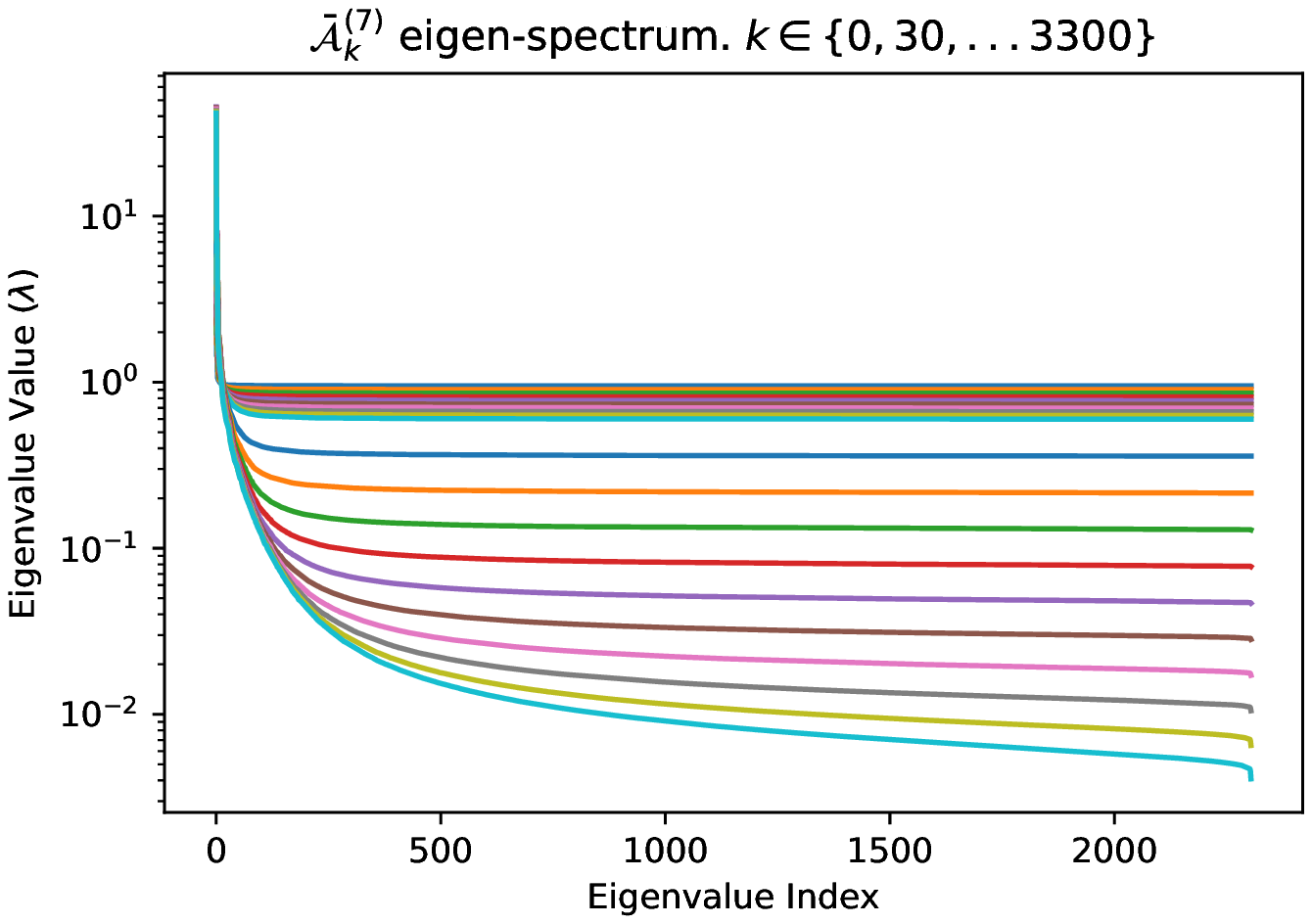}
	\includegraphics[trim={0.1cm 0.1cm 1.5cm 0.4cm},clip,width=0.323\textwidth]{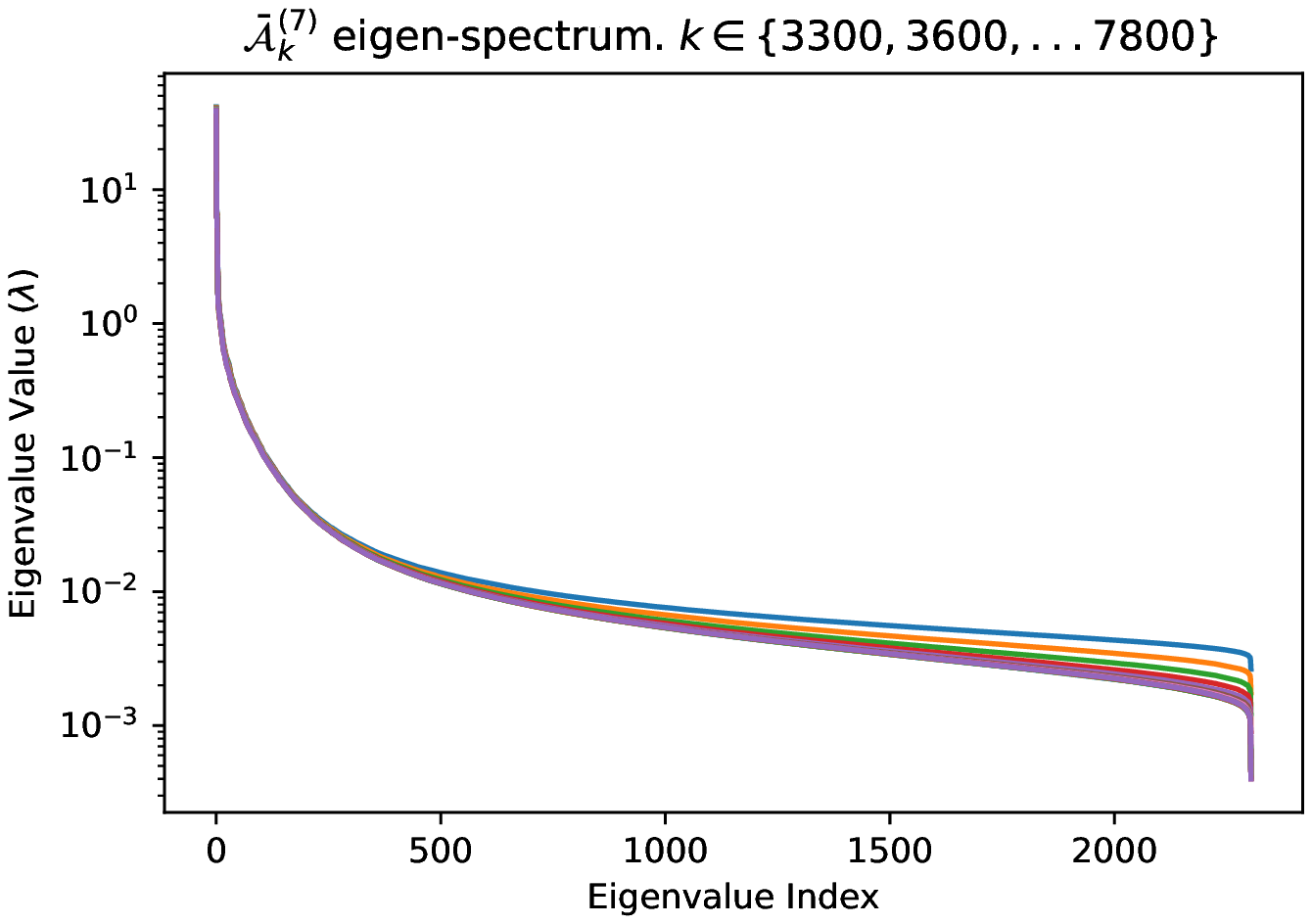}
	\includegraphics[trim={0.1cm 0.1cm 1.5cm 0.4cm},clip,width=0.323\textwidth]{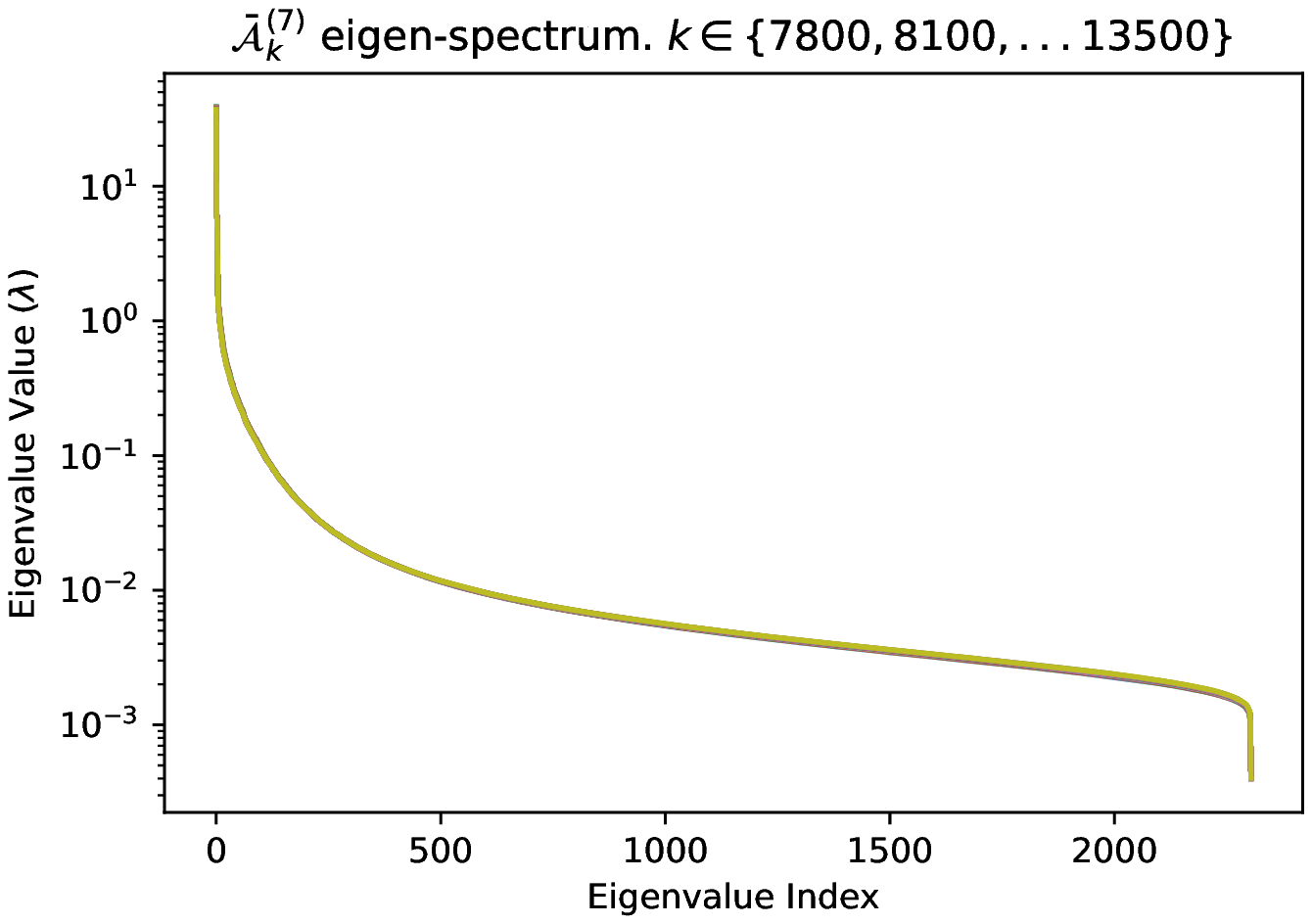}
	
	\includegraphics[trim={0.1cm 0.1cm 1.5cm 0.4cm},clip,width=0.323\textwidth]{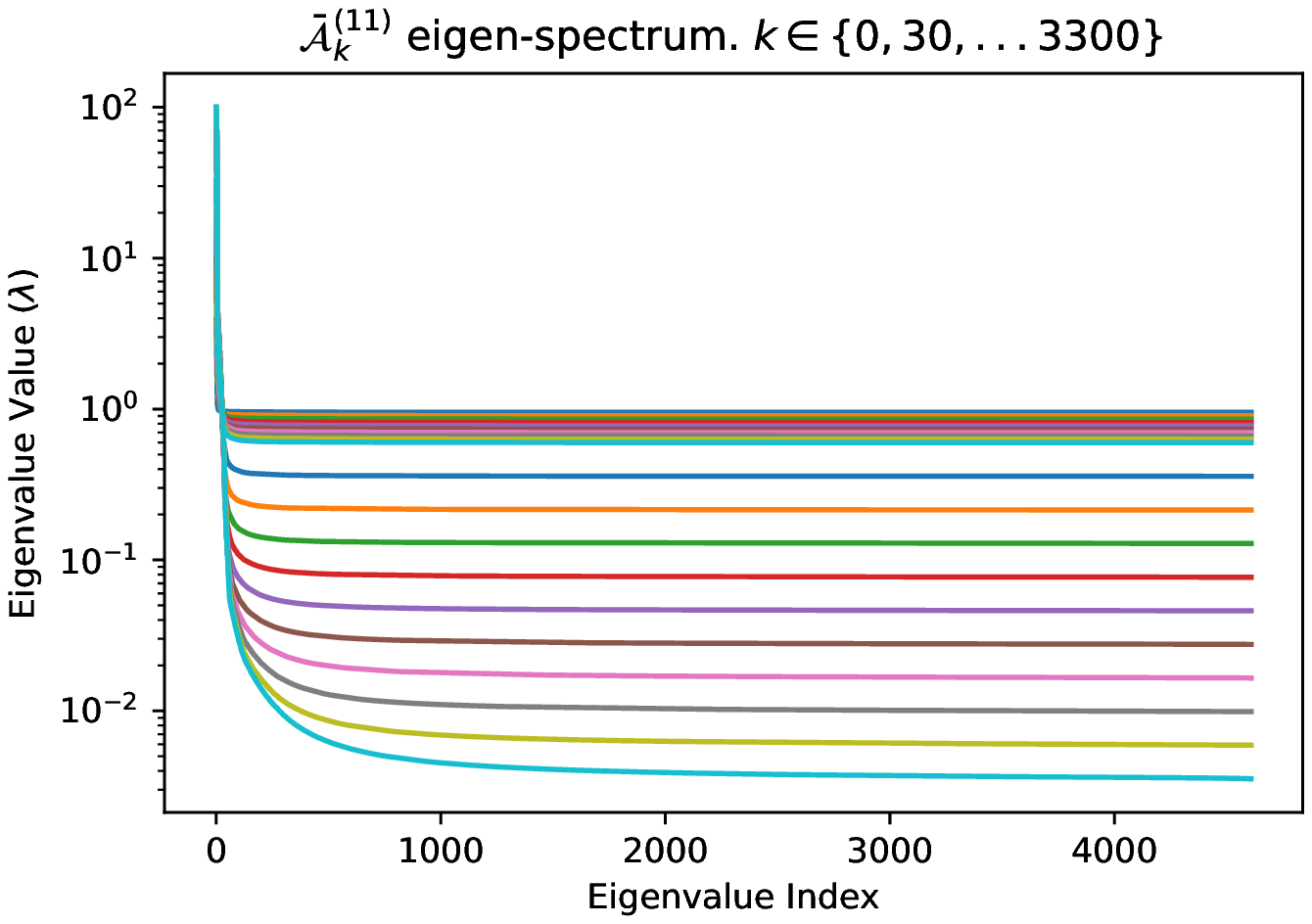}
	\includegraphics[trim={0.1cm 0.1cm 1.5cm 0.4cm},clip,width=0.323\textwidth]{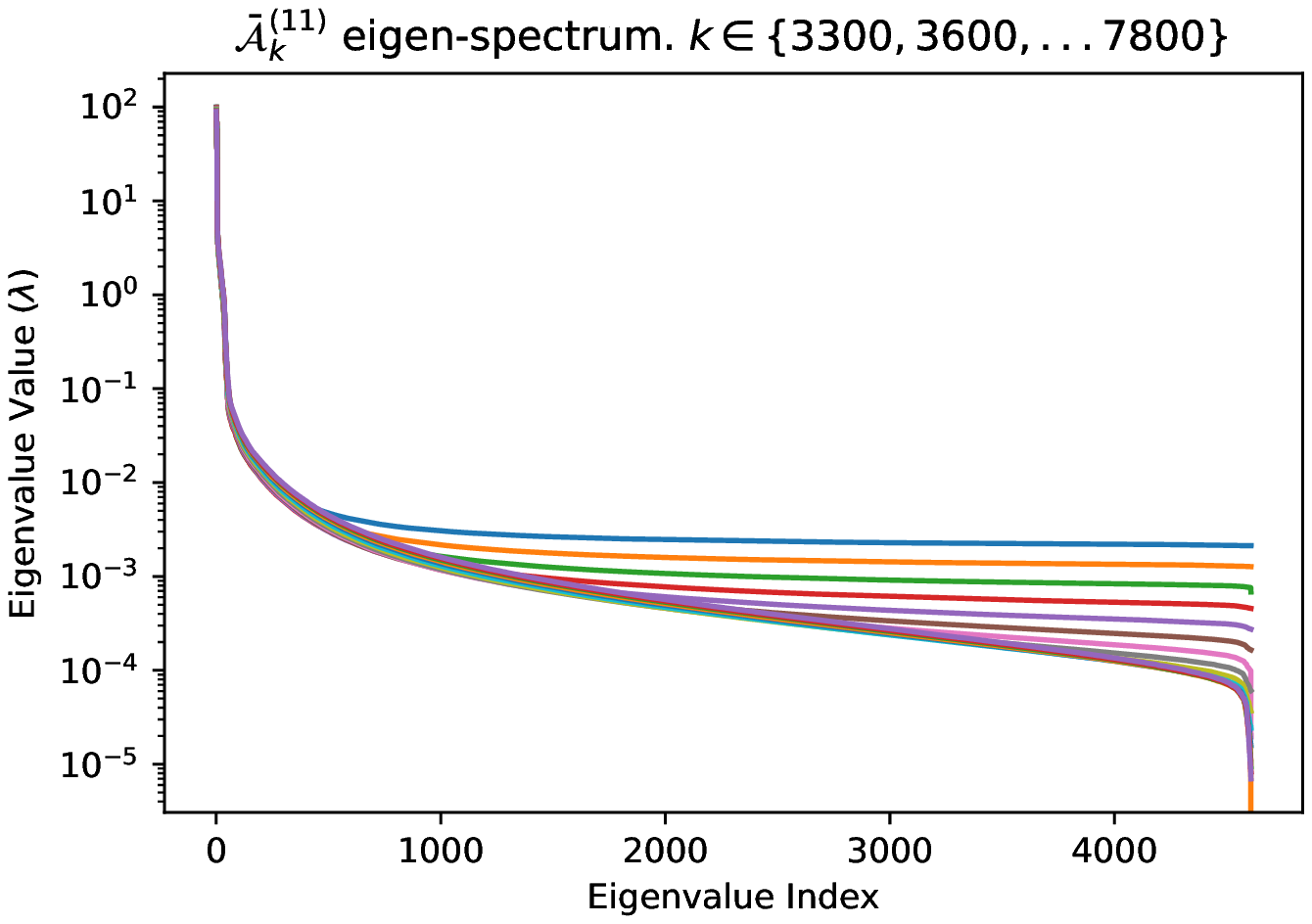}
	\includegraphics[trim={0.1cm 0.1cm 1.5cm 0.4cm},clip,width=0.323\textwidth]{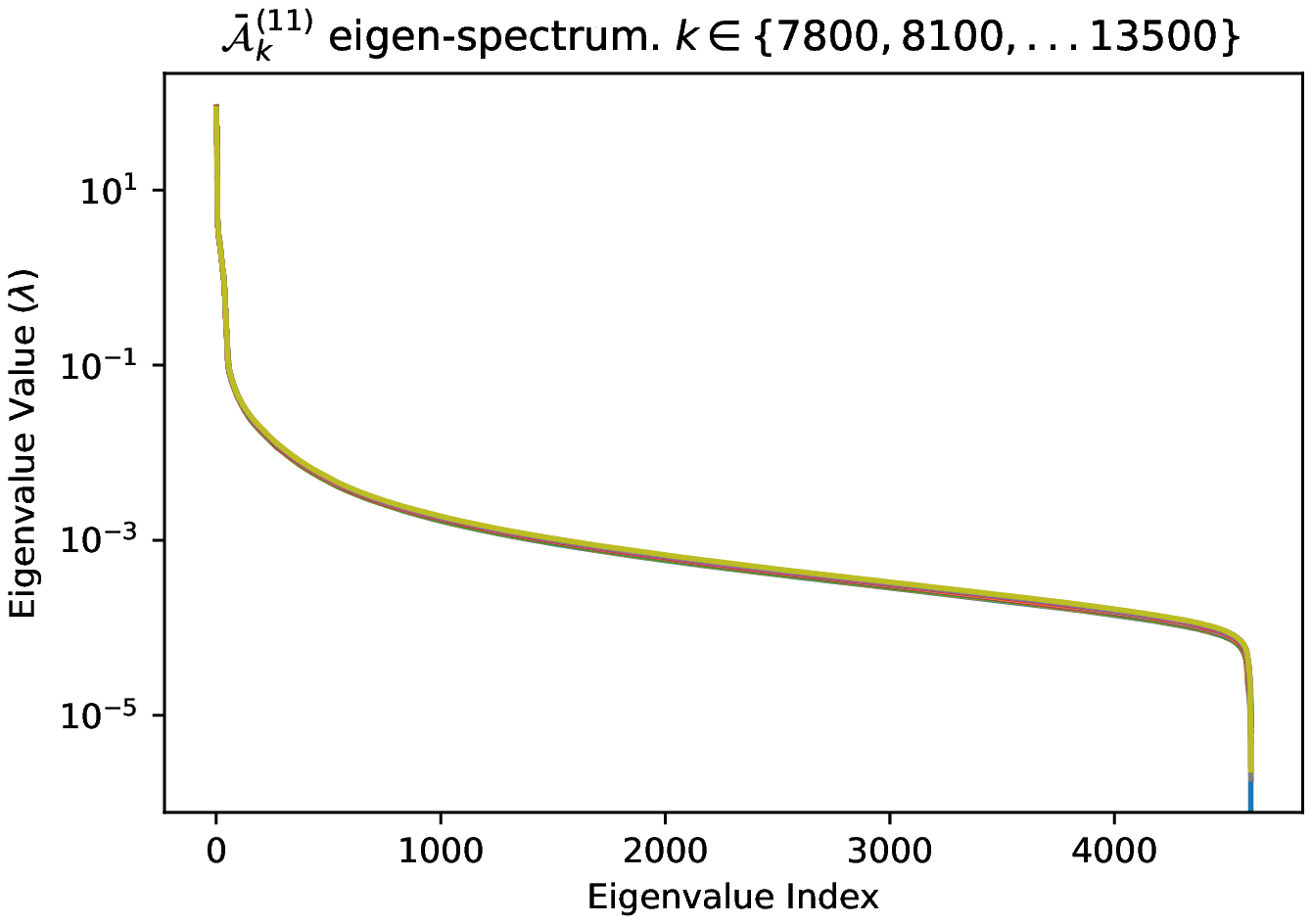}
	
	\includegraphics[trim={0.1cm 0.1cm 1.5cm 0.4cm},clip,width=0.323\textwidth]{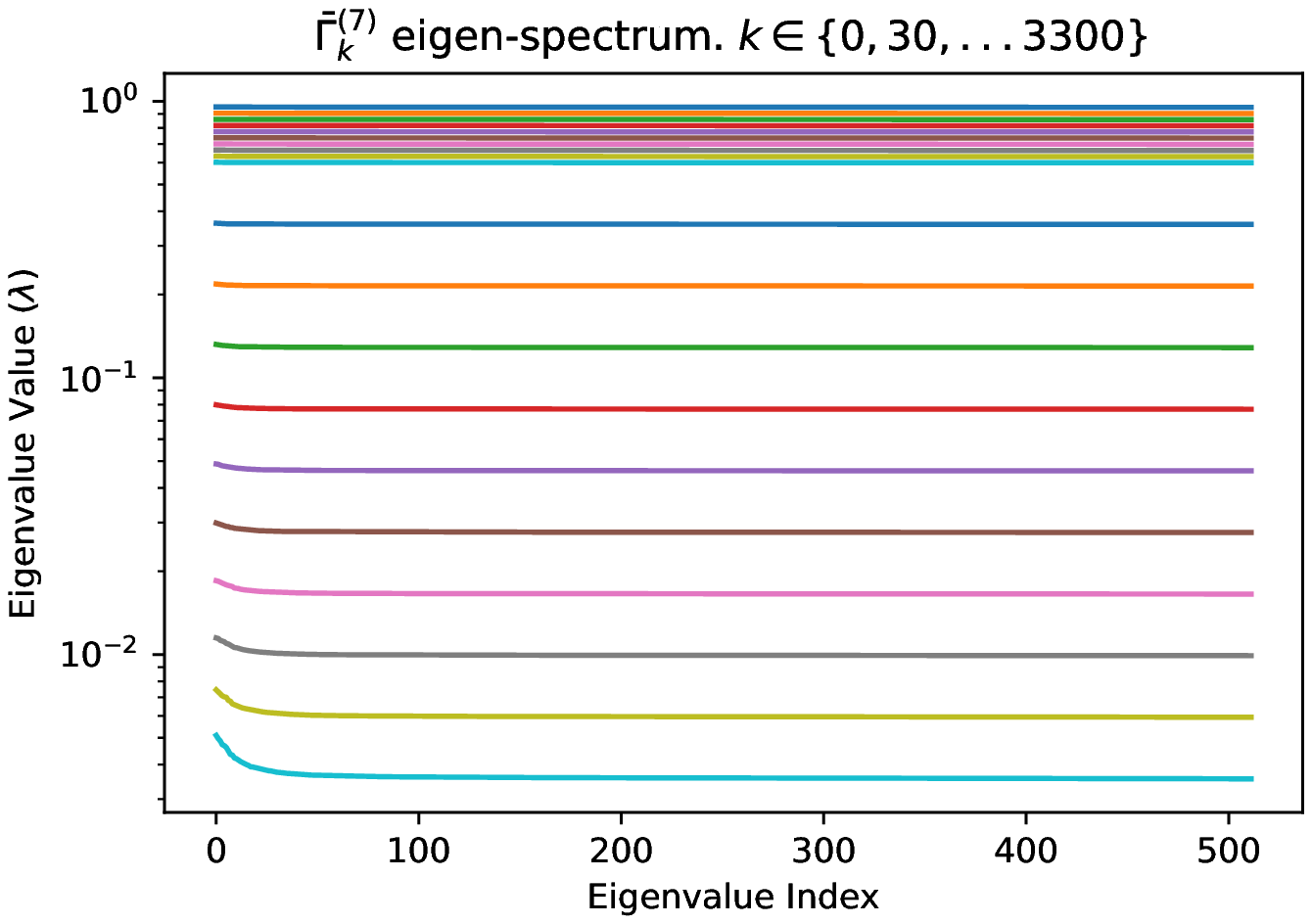}
	\includegraphics[trim={0.1cm 0.1cm 1.5cm 0.4cm},clip,width=0.323\textwidth]{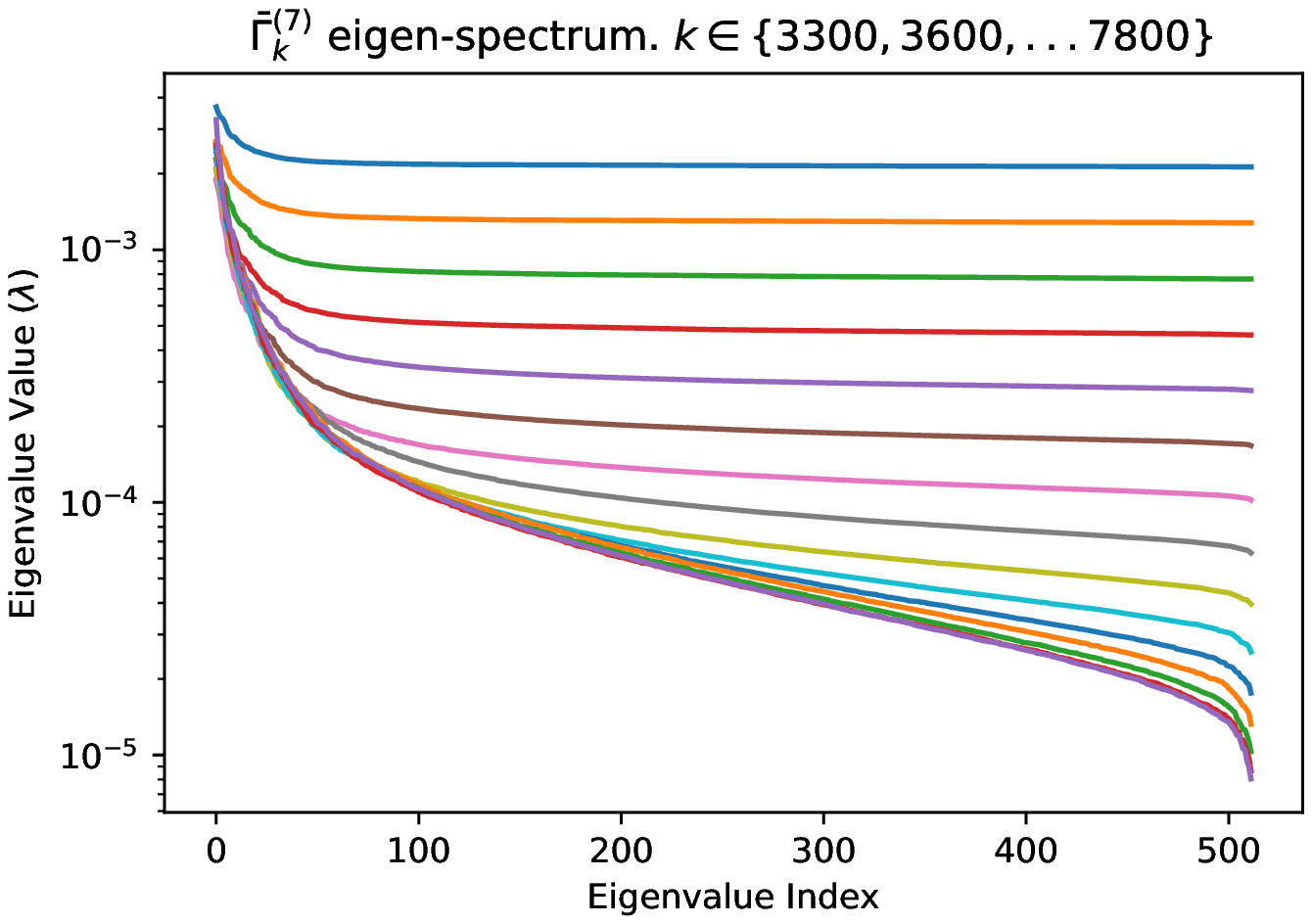}
	\includegraphics[trim={0.1cm 0.1cm 1.5cm 0.4cm},clip,width=0.323\textwidth]{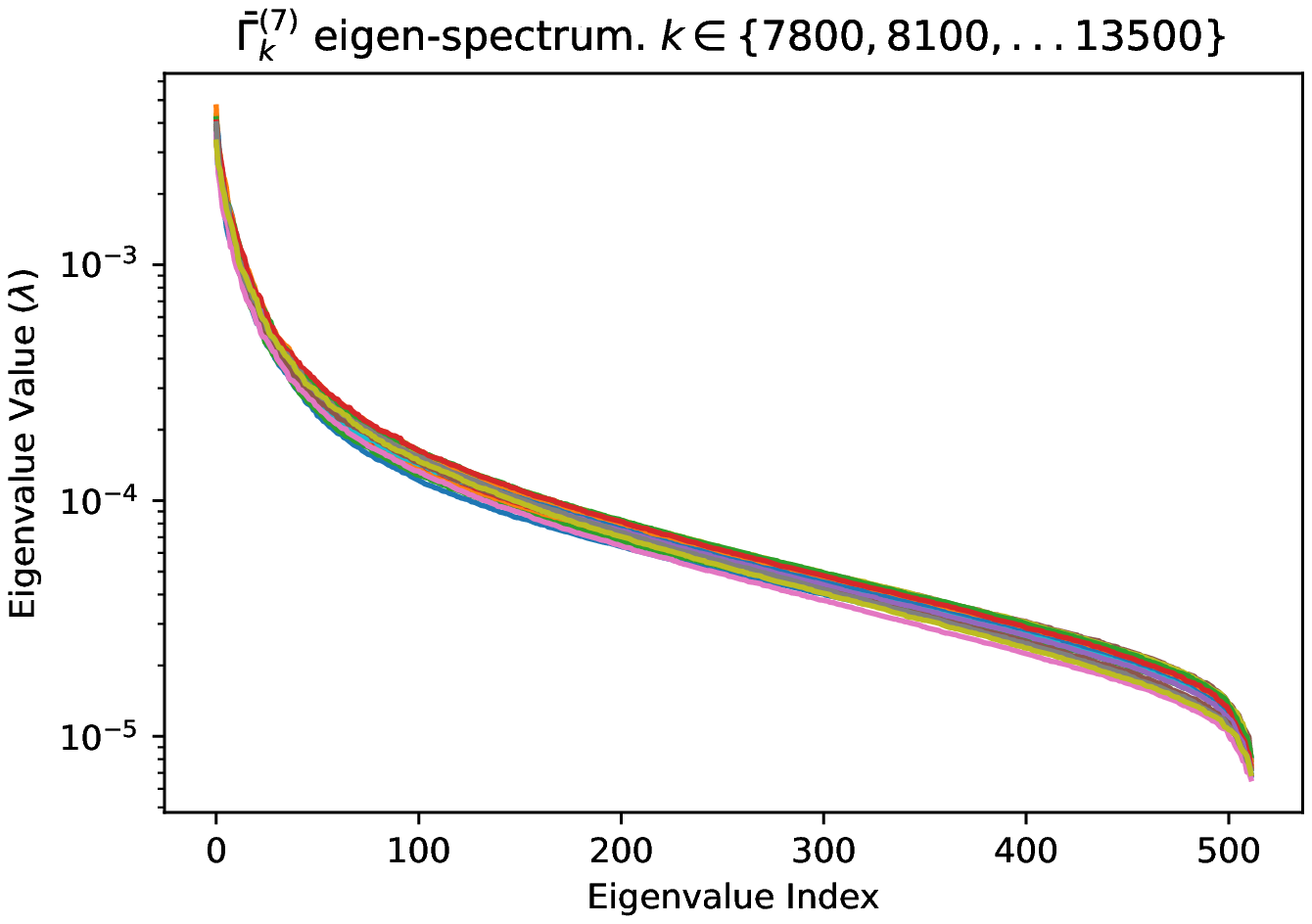}
	
	\includegraphics[trim={0.1cm 0.1cm 1.5cm 0.4cm},clip,width=0.323\textwidth]{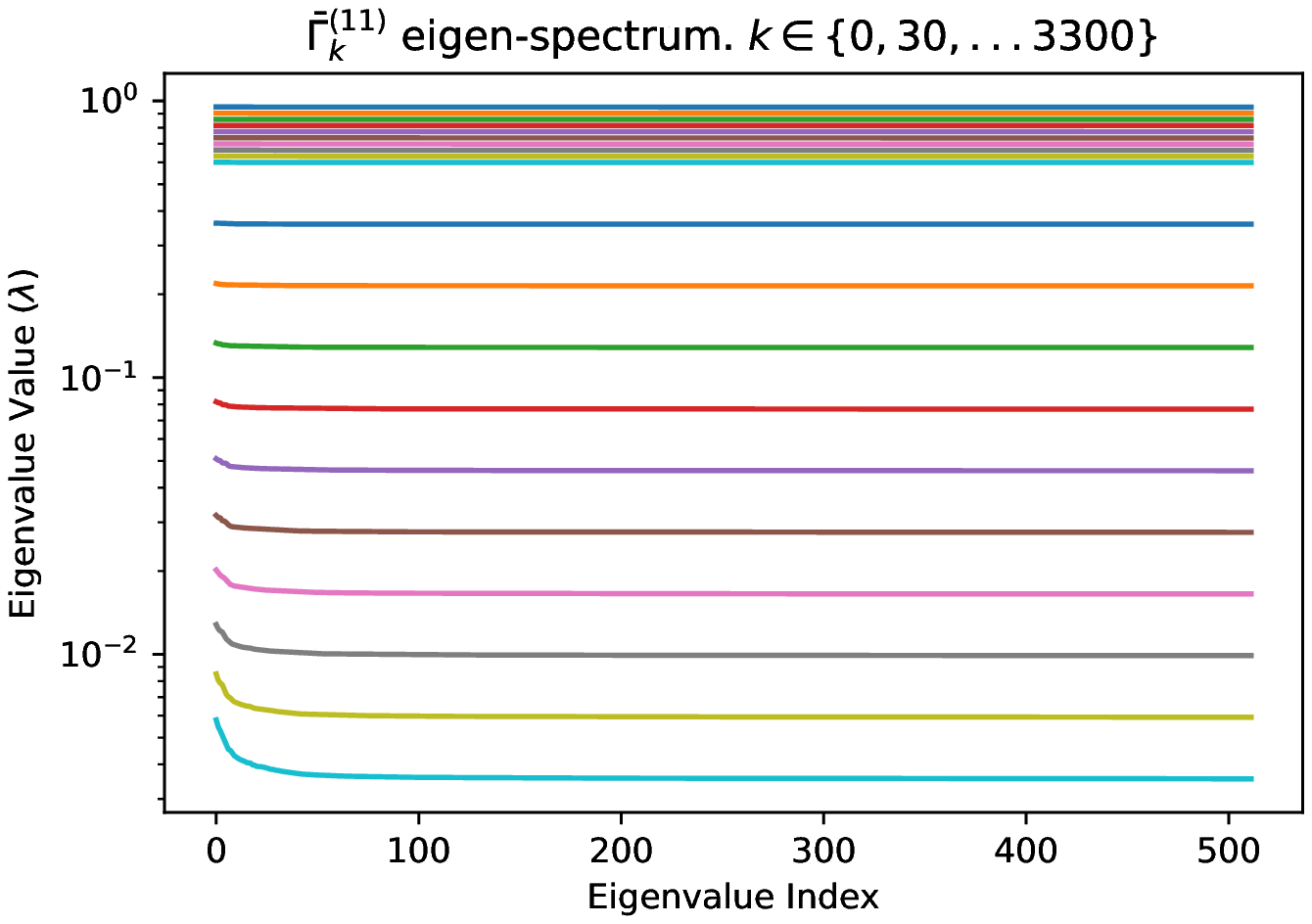}
	\includegraphics[trim={0.1cm 0.1cm 1.5cm 0.4cm},clip,width=0.323\textwidth]{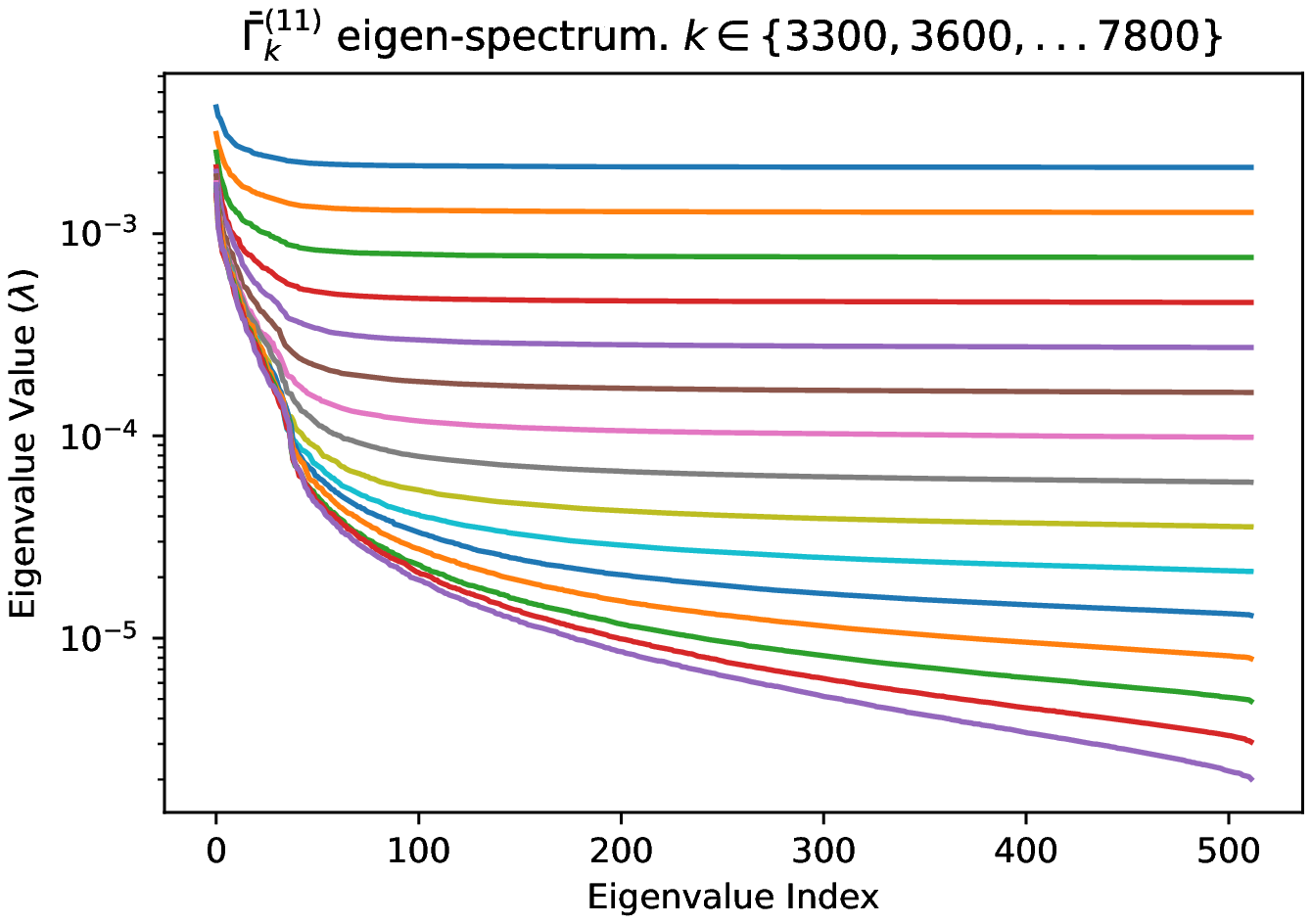}
	\includegraphics[trim={0.1cm 0.1cm 1.5cm 0.4cm},clip,width=0.323\textwidth]{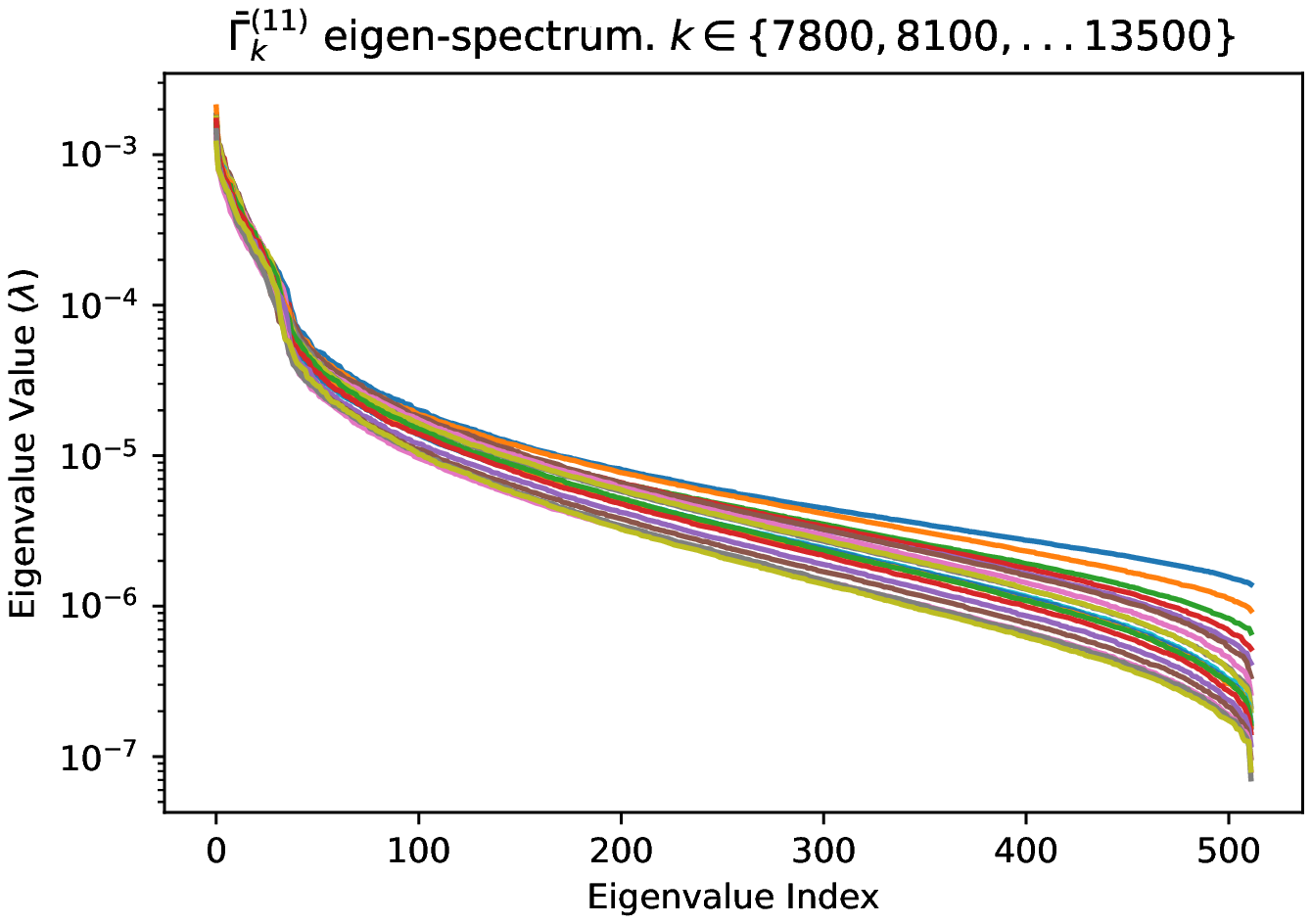}
	\caption{K-Factors eigen-spectrum: layers 7 and 11 of  VGG16\_bn for CIFAR10 dataset. Each curve represents the spectrum for a specific step $k$. } 
	\label{eigenspectrum_results_joint_acc_Loss_2}
\end{figure}

\subsubsection{Numerical Investigation of K-Factors Eigen-Spectrum}
We ran \textsc{k-fac} for 70 epochs, with the specifications outlined in \textit{Section 5} (but with $T_{K,U}=T_{K,I}=30$). We saved the eigen-spectrum every 30 steps if $k<300$, and every 300 steps otherwise. Only results for layers 7 and 11 are shown for the sake of brevity, but they were virtually identical for all other layers. We see that for low $k$, all eigenvalues are close to unity, which is due to $\bar{\mathcal A}$ and $\bar \Gamma$ being initialized to the identity. However, the spectrum rapidly develops a strong decay (where more than 1.5 orders of magnitude are decayed within the first 200 eigenvalues). It takes $\bar {\mathcal A}$ about 500 steps (that is about 2.5 epochs) and $\bar \Gamma$ about 5100 steps (26 epochs) to develop this strong spectrum decay. We consider 1.5 orders of magnitude a strong decay because the K-Factors regularization that we found to work best is around $\lambda_{\text{max}}/10$ (for which any eigenvalue below $\lambda_{\text{max}}/33$ can be considered zero without much accuracy loss). Thus, truncating our K-Factors to an $r\approx220$ worked well in practice. Importantly, once the spectrum reaches its equilibrium state, we get this 1.5 orders of magnitude decay within 200 modes irrespective of the size of the K-factor ($d_M$). This aligns with the intuition provided by \textit{Proposition 3.1}.

\section{Speeding Up EA K-Factors Inversion}
We now  present two approaches for speeding up \textsc{k-fac}, which avoid the typically used \textsc{evd} of the K-factors through obtaining approximations to the low-rank truncations of these \textsc{evd}s. The ideas are similar in spirit and presented in the order of increasing computational saving (and reducing accuracy).
\subsection{Proposed Optimizer: RSVD K-FAC (RS-KFAC)}
Instead of computing the eigen-decompositions of the EA-matrices (K-Factors) $\bar {\mathcal A}$ and $\bar \Gamma$ (in \textit{line 12} of \textit{Algorthm 1}; of time complexity $\mathcal O(d_{\mathcal A}^3)$ and $\mathcal O(d_\Gamma^3)$), we could settle for using a rank $r$ \textsc{rsvd}
approximation:
\begin{equation}
\bar {\mathcal A} \,\RSVDapprox\, \tilde U_{A}\tilde D_{A}\tilde U_{A}^T,\,\,\, \text{and }\,\,\,
\bar \Gamma\, \RSVDapprox \,  \tilde U_{\Gamma}\tilde D_{\Gamma}\tilde U_{\Gamma}^T,
\end{equation}
where $\tilde U_{A}\in \mathbb R^{d_{\mathcal A} \times r}$,  $\tilde U_{\Gamma}\in \mathbb R^{d_\Gamma \times r}$,  and $\tilde D_{A}, \tilde D_{\Gamma}\in \mathbb R^{r \times r}$.

Using this trick, we reduce the computation cost of \textit{line 12} in \textit{Algorithm 1} from $\mathcal O(d_{\mathcal A}^3 + d_\Gamma^3)$ to $\mathcal O((d_{\mathcal A}^2 +  d_\Gamma^2 ) (r+r_l))$ when using an oversampling parameter $r_l$. This is a dramatic reduction since we can choose $(r+r_l)\ll \min(d_{\mathcal A},d_\Gamma)$ with minimal truncation error, as we have seen in \textit{Section 3}. As discussed in \textit{Section 2.3}, for \textsc{rsvd} the projection error is virtually zero, and thus small truncation error means our \textsc{rsvd} approach will give very close results to using the full eigenspectrum.
Once we have the approximate low-rank truncations, we estimate
\begin{equation}
(\bar \Gamma + \lambda I)^{-1} V \approx (\tilde U_{\Gamma,r} \tilde D_{\Gamma,r} \tilde U_{\Gamma,r}^T + \lambda I)^{-1} V,
\end{equation}
where $\lambda$ is the regularization parameter (applied to K-factors), and then compute
\begin{equation}
\begin{split}
(\tilde U_{\Gamma,r} \tilde D_{\Gamma,r} \tilde U_{\Gamma,r}^T  + \lambda I)^{-1} V = \tilde U_{\Gamma, r}\biggl[( \tilde D_{\Gamma,r} + \lambda I)^{-1} - \frac{1}{\lambda}I\biggr] \tilde U_{\Gamma,r}^T V + \frac{1}{\lambda}V.
\end{split}
\label{inversion_from_RSVD}
\end{equation}
We use (\ref{inversion_from_RSVD}) because its r.h.s. is cheaper to compute than its l.h.s. Note that computing (\ref{inversion_from_RSVD}) has complexity $\mathcal O(rd_{\Gamma} + 2rd_{\Gamma}^2)$, which is better than computing \textit{line 15} of \textit{Algorithm 1} of complexity $\mathcal O(d_\Gamma^3)$. We take a perfectly analogous approach for $V(\bar {\mathcal A}+\lambda I)^{-1}$. The \textsc{rs-kfac} algorithm is obtained by replacing \textit{lines 10 - 15} in \textit{Algorithm 1} with the for loop shown in \textit{Algorithm 4}. Over-all \textsc{rs-kfac} scales like $\mathcal O(d_M^2(r+r_l))$ (setting $d_M=d_{\mathcal A}^{(l)}=d_{\Gamma}^{(l)}$, $\forall l$ for simplicity).

\begin{algorithm}[H]
	\footnotesize
	\label{RS_KFAC_algorithm}
	\caption{\textsc{rs-kfac} (our first proposed algorithm)}
	Replace \textit{lines 10 - 15} in \textit{Algorithm 1} with:

	\For{$l=0,1,...,N_L$}
	{
		\tcp{Get RSVD of $\bar {\mathcal A}^{(l)}_k$ and $\bar \Gamma^{(l)}_k$ for inverse application}
		$\tilde U^{(l)}_{A,k}\tilde D^{(l)}_{A,k} (\tilde V^{(l)}_{A,k})^T = \text{RSVD}(\bar {\mathcal A}^{(l)}_k)$;	$\tilde U_{\Gamma,k}^{(l)}\tilde D_{\Gamma,k}^{(l)} (\tilde V^{(l)}_{\Gamma,k})^T = \text{RSVD}(\bar \Gamma^{(l)}_k)$
		
		\tcp{Use RSVD factors to approx.\ apply inverse of K-FAC matrices}
		
		$J^{(l)}_k = \text{Mat}(g_k^{(l)})$
		
		$M^{(l)}_k = J^{(l)}_k \tilde V^{(l)}_{A, k}\big[( \tilde D^{(l)}_{A,k} + \lambda I)^{-1} - \frac{1}{\lambda}I\big] (\tilde V^{(l)}_{A,k})^T+ \frac{1}{\lambda}J^{(l)}_k$
		
		$S^{(l)}_k = \tilde V^{(l)}_{\Gamma, k}\biggl[( \tilde D^{(l)}_{\Gamma,k} + \lambda I)^{-1} - \frac{1}{\lambda}I\biggr] (\tilde V^{(l)}_{\Gamma,k})^T M^{(l)}_k + \frac{1}{\lambda}M^{(l)}_k $
		
		$s^{(l)}_k = \text{vec}(S^{(l)}_k)$
	}
	
\end{algorithm}
Note that the RSVD subroutine in \textit{line 4} of \textit{Algorithm 4} may be executed using the \textsc{rsvd} in \textit{Algorithm 2}, but using different \textsc{rsvd} implementations would not significantly change our discussion. As we have discussed in \textit{Section 2.3.1}, even though $\tilde U^{(l)}_{A,k}$ should equal $\tilde V^{(l)}_{A,k}$ since $\bar {\mathcal A}_k^{(l)}$ is square s.p.s.d., the \textsc{rsvd} algorithm returns two (somewhat) different matrices, of which the more accurate one is the ``V-matrix''. The same observation also applies to $\Gamma$-related quantities.

\subsection{Proposed Optimizer: SREVD K-FAC (SRE-KFAC)}
Instead of using \textsc{rsvd} in \textit{line 4} of \textit{Algorithm \ref{RS_KFAC_algorithm}}, we can exploit the symmetry and use \textsc{srevd} (e.g.\ with \textit{Algorithm \ref{Symmetric_randomized_EVD}}). This would reduce the computation cost of that line by a constant factor, altough the computational complexity would be the same: $\mathcal O\big((d_{\mathcal A}^2 + d_{\Gamma}^2)(r+r_l)\big)$. However, this cost reduction comes at the expense of reduced accuracy, because \textsc{srevd} has significant \textit{projection error} (unlike \textsc{rsvd}; recall \textit{Section 2.3}). We refer to this algorithm as \textsc{sre-kfac} and briefly present it in \textit{Algorithm 5}. Note that in \textit{line 4} of \textit{Algorithm 5} we assign $\tilde V\leftarrow \tilde U$ to avoid rewriting \textit{lines 7-8} of \textit{Algorithm 4} with $\tilde V$'s replaced by $\tilde U's$. Over-all \textsc{sre-kfac} scales like $\mathcal O(d_M^2(r+r_l))$ (setting $d_M=d_{\mathcal A}^{(l)}=d_{\Gamma}^{(l)}$, $\forall l$ for simplicity of exposition).

\begin{algorithm}[H]
	\caption{\textsc{sre-kfac} (our second proposed algorithm)}
	Replace lines \textit{lines 3 - 4 in Algorithm 4} with:
	
	\tcp{Get SREVD of $\bar {\mathcal A}^{(l)}_k$ and $\bar \Gamma^{(l)}_k$ for inverse application}
	
	$\tilde U^{(l)}_{A,k}\tilde D^{(l)}_{A,k} (\tilde U^{(l)}_{A,k})^T = \text{SREVD}(\bar {\mathcal A}^{(l)}_k)$; $\tilde U_{\Gamma,k}^{(l)}\tilde D_{\Gamma,k}^{(l)} (\tilde U^{(l)}_{\Gamma,k})^T = \text{SREVD}(\bar \Gamma^{(l)}_k)$
	
	$\tilde V^{(l)}_{A,k} = \tilde U^{(l)}_{A,k}$; 	$\tilde V^{(l)}_{\Gamma,k} = \tilde U^{(l)}_{\Gamma,k}$
\end{algorithm}

\subsection{Direct Idea Transfer to Other Applications}
\textbf{Application to \textsc{ek-fac}:} We can apply the method directly to \textsc{ek-fac} (a \textsc{k-fac} improvement; \cite{EKFAC}) as well.

\textbf{Application to \textsc{kld-wrm} algorithms:}
Our idea can be directly applied to the \textsc{kld-wrm} family (see \cite{kld_wrm}) when \textsc{k-fac} is used as an implementation \textit{``platform''}. Having a smaller optimal $\rho$ ($0.5$ as opposed to $0.95$), \textsc{kld-wrm} instantiations may benefit more from our porposed ideas, as they are able to use even lower target-ranks in the \textsc{rsvd} (or \textsc{srevd}) for the same desired accuracy. To see this, consider setting $\rho := 0.5$ (instead of $\rho = 0.95$) in the practical calculation underneath \textit{Proposition 3.1}. Doing so reduces the required number of retained eigenvalues down to $2304$ from $29184$.



\subsection{Partly Closing the Complexity Gap between K-FAC and SENG}
It is important to realise that this section gives us more than a way of significantly speeding K-FAC for large net widths (at negligible accuracy loss). It tells us that (based on the Discussion in \textit{Section 3} and \textit{Proposition 3.1}) the scaling of $\mathcal O(d_M^3)$ with layer width is \textit{not} inherent to K-FAC (at least not when $d_M\gg n_{\text{BS}}$), and that we can obtain scaling of $\mathcal O(d_M^2)$ for K-FAC at practically no accuracy loss. 

This opportunity conceptually arises in a simple way. Roughly speaking, we have much less information in the K-factor estimate (scales with $n_{\text{BS}}$; and we cannot take too large batch-sizes) than would be required to estimate it accurately given its size $d_M\times d_M$ (when $d_M\gg n_{\text{BS}}$). Thus, whether the true K-factor has strong eigen-spectrum decay or not does not matter, our EA estimates are \textit{bound} to exhibit it. So what causes a problem actually solves another: we cannot accurately estimate the K-factors for large $d_M$ given our bacth-size limitation - but this puts us in a place where our approximate decomposition/inversion computations which scale like $\mathcal O(d_M^2)$ are virtually as good as the exact methods which scale like $\mathcal O(d_M^3)$. 

This brings K-FAC practically closer to the computational scaling of SENG\footnote{See \textit{Section 3.3.2} or the original paper (\cite{SENG}) for details.} (the more succesful practical NG implementation) To see this, note that we have $\mathcal O(d_M^3)$ for K-FAC, $\mathcal O(d_M^2)$ for \textit{Randomized K-FACs}, and $\mathcal O(d_M)$ for SENG. Conceptually, SENG has better scaling as it exploits this lack of information to speed-up computation by removing unnecesary ones. We hereby in this paper implicitly show that we can do a similar thing for K-FAC and obtain a better scaling with $d_M$!

\section{Numerical Results: Proposed Algorithms Performance}
We now numerically compare \textsc{rs-kfac} and \textsc{sre-kfac} with \textsc{k-fac} (the baseline we improve upon) and \textsc{seng} (another NG implementation which typically outperforms \textsc{k-fac}; see \cite{SENG}). We did not test \textsc{sgd}, as this underpeforms \textsc{seng} (see \textit{Table 4} in \cite{SENG}). We consider the CIFAR10 dataset with a modified\footnote{We add a 512-in 512-out FC layer with dropout ($p=0.5$) before the final FC layer.} version of batch-normalized VGG16 (VGG16\_bn). All experiments ran on a single \textit{NVIDIA Tesla V100-SXM2-16GB} GPU. The accuracy we refer to is always \textit{test accuracy}.

\subsubsection{Implementation Details} 
For \textsc{seng}, we used the implementation from the \textit{official github repo} with the hyperparameters\footnote{\textbf{Repo:} https://github.com/yangorwell/SENG. \textbf{Hyper-parameters:} \textit{label\_smoothing = 0, fim\_col\_sample\_size = 128, lr\_scheme = 'exp', lr = 0.05, lr\_decay\_rate = 6, lr\_decay\_epoch = 75, damping = 2, weight\_decay = 1e-2, momentum = 0.9, curvature\_update\_freq = 200. Omitted params.\ are default.} } directly recommended by the authors for the problem at hand (via email). \textsc{k-fac} was slightly adapted  from \textit{alecwangcq's github}\footnote{\textbf{Repo: }https://github.com/alecwangcq/KFAC-Pytorch}. Our proposed solvers were built on that code. For \textsc{k-fac}, \textsc{rs-kfac} and \textsc{sre-kfac} we performed manual tuning. We found that no momentum, weight\_decay = 7e-04,  $T_{K,U} = 10$, and $\rho = 0.95$, alongside with the schedules $T_{K,I}(n_{\text{ce}}) = 50 - 20 \mathbb I_{n_{\text{ce}}\geq 20}$, $\lambda_K(n_{\text{ce}}) = 0.1 - 0.05\mathbb I_{n_{\text{ce}}\geq 25} - 0.04\mathbb I_{n_{\text{ce}}\geq 35}$, $\alpha_k(n_{ce}) = 0.3 - 0.1\mathbb I_{n_{ce}\geq 2} - 0.1\mathbb I_{n_{ce}\geq 3} - 0.07\mathbb I_{n_{ce}\geq 13} - 0.02\mathbb I_{n_{ce}\geq 18} - 0.007\mathbb I_{n_{ce}\geq 27} - 0.002\mathbb I_{n_{ce}\geq 40}$ (where $n_{\text{ce}}$ is the number of the current epoch) worked best for all three \textsc{k-fac} based solvers. The hyperparameters specific to \textsc{rs-kfac} and \textsc{sre-kfac} were set to $n_{\text{pwr-it}}= 4$, $r(n_{\text{ce}}) = 220 + 10\mathbb I_{n_{\text{ce}}\geq 15}$, $r_l(n_{\text{ce}}) = 10 + \mathbb I_{n_{ce}\geq 22}  + \mathbb I_{n_{\text{ce}}\geq 30}$. We set $n_{\text{BS}}= 256$ throughout. We implemented all our \textsc{k-fac}-based algorithms in the empirical NG spirit (using $y$ from the given labels when computing the backward K-factors rather than drawing $y\sim p (y|h_\theta(x))$; see \cite{New_insights_and_perspectives} for details). We performed 10 runs of 50 epochs for each $\{$solver, batch-size$\}$ pair\footnote{\textbf{Our codes repo:} https://github.com/ConstantinPuiu/Randomized-KFACs}.
\begin{table}[t]
	\caption{CIFAR10 VGG16\_bn results summary. All solvers reached $91.5\%$ accuracy within the allocated 50 epochs for 10 out of 10 runs. Some solvers did not reach $92\%$ accuracy on all of their 10 runs, and this is shown in the sixth column of the table. Columns 2-4 show the time to get to a specific test accuracy. The fifth column shows time per epoch. All times are in seconds and presented in the form: mean $\pm$ standard deviation. For time per epoch, statistics are obtained across 500 samples (50 epochs $\times$ 10 runs). For times to a specific accuracy, statistics are obtained based only on the runs where the solver indeed reached the target accuracy (eg.\ for $92\%$, the 5 successful runs are used for \textsc{k-fac}). The last column of the table shows number of epochs to get to $92\%$ accuracy (results format is analogous to the ones of column 4).}
	
	
	
	\label{MNIST_results_table}
	\centering
	\begin{tabular}{|c|c|c|c|c|c|c|c|c|c|}
		\hline
		\makecell{} & \makecell{ $t_{acc\geq90\%}$} &\makecell{$t_{acc\geq91.5\%}$} & \makecell{$t_{acc\geq92\%}$} &  \makecell{$t_{\text{epoch}}$}  &  \makecell{Runs hit $92\%$} & \makecell{$\mathcal N_{\text{acc}\geq 92\%}$} \\
		\hline
		
		\hline
		
		\hline
		
		\textsc{seng} & $673.6\pm34.4$ & $693.2\pm28.2$ & $718.1\pm26.0$ & $16.6\pm0.4$ & 10 out of 10 & $43.3\pm0.9$ \\
		\hline
		
		\textsc{k-fac} & $1449 \pm 8.7$ & $1971 \pm 225$ & $2680 \pm 636$ & $75.5 \pm 3.4$ & 5 out of 10 & $35.4\pm8.3$\\
		\hline
		
		
		\textsc{rs-kfac} & $445.8\pm10.9$ & $600.7\pm4.9$ & $732.6\pm153.1$ & $32.6\pm0.9$ &  10 out of 10 & $23.0\pm4.7$\\
		\hline
		
		\textsc{sre-kfac}  \ & $439.4\pm28.5$ & $582.2\pm24.1$ & $785.3\pm155.6$ & $30.0\pm0.4$ & 7 out of 10 & $26.3\pm5.1$ \\
		
		\hline
	\end{tabular}	
\end{table}
\subsubsection{Results Discussion} 
\textit{Table 1} shows important summary statistics. We see that the time per epoch is $\approx 2.4\times$ lower for our solvers than for \textsc{k-fac}. This was expected given we reduce time complexity from cubic to quadratic in layer width! In accordance with our discussion in \textit{Section 4.2}, we see that \textsc{sre-kfac} is slightly faster per epoch than \textsc{rs-kfac}. Surprisingly, we see that the number of epochs to a target accuracy (at least for $92\%$) is also smaller for \textsc{rs-kfac} and \textsc{sre-kfac} than for \textsc{k-fac}. This indicates that dropping the low-eigenvalue modes does not seem to hinder optimization progress, but provide a further benefit instead. As a result, the time to a specific target accuracy is improved by a factor of $3$ - $4\times$ when using \textsc{rs-kfac} or \textsc{sre-fac} as opposed to \textsc{k-fac}. Note that \textsc{sre-kfac} takes more epochs to reach a target accuracy than \textsc{rs-kfac}. This is due \textsc{sre-kfac} further introducing a \textit{projection error} compared to \textsc{rs-kfac} (see \textit{Section 4.2}). For the same reason, \textsc{rs-kfac} always achieves $92\%$ test accuracy while \textsc{sre-kfac} only does so 7 out of 10 times. Surprisingly, \textsc{k-fac} reached $92\%$ even fewer times. We believe this problem appeared in \textsc{k-fac} based solvers due to a tendency to overfit, as can be seen in \textit{Figure 2}. 
\enlargethispage{5ex}

When comparing to \textsc{seng}, we see that our proposed \textsc{k-fac} improvements perform slightly better for $91\%$ and $91.5\%$ target test accuracy, but slightly worse for $92\%$. We believe this problem will vanish if we can fix the over-fit of our \textsc{k-fac} based solvers. Overall, the numerical results show that our proposed speedups give substantially better implementations of \textsc{k-fac}, with time-to-accuracy speed-up factors of $\approx3.3\times$. \textit{Figure 2} shows an in-depth view of our results.


\begin{figure}[t]
	\centering

	\includegraphics[trim={0.1cm 0.15cm 1.6cm 0.9cm},clip,width=0.2935\textwidth]{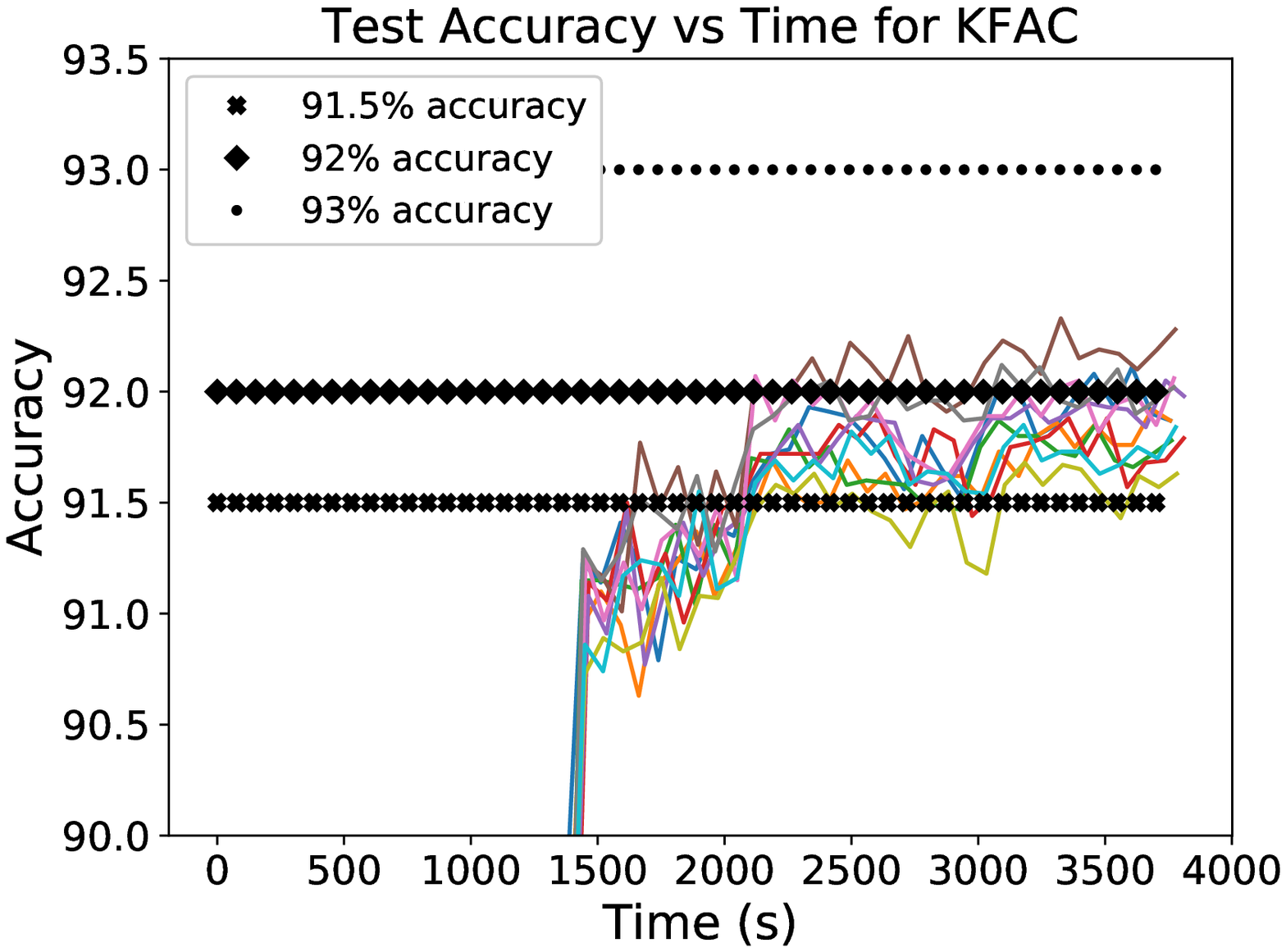}
	\includegraphics[trim={0.1cm 0.15cm 1.6cm 0.9cm},clip,width=0.2935\textwidth]{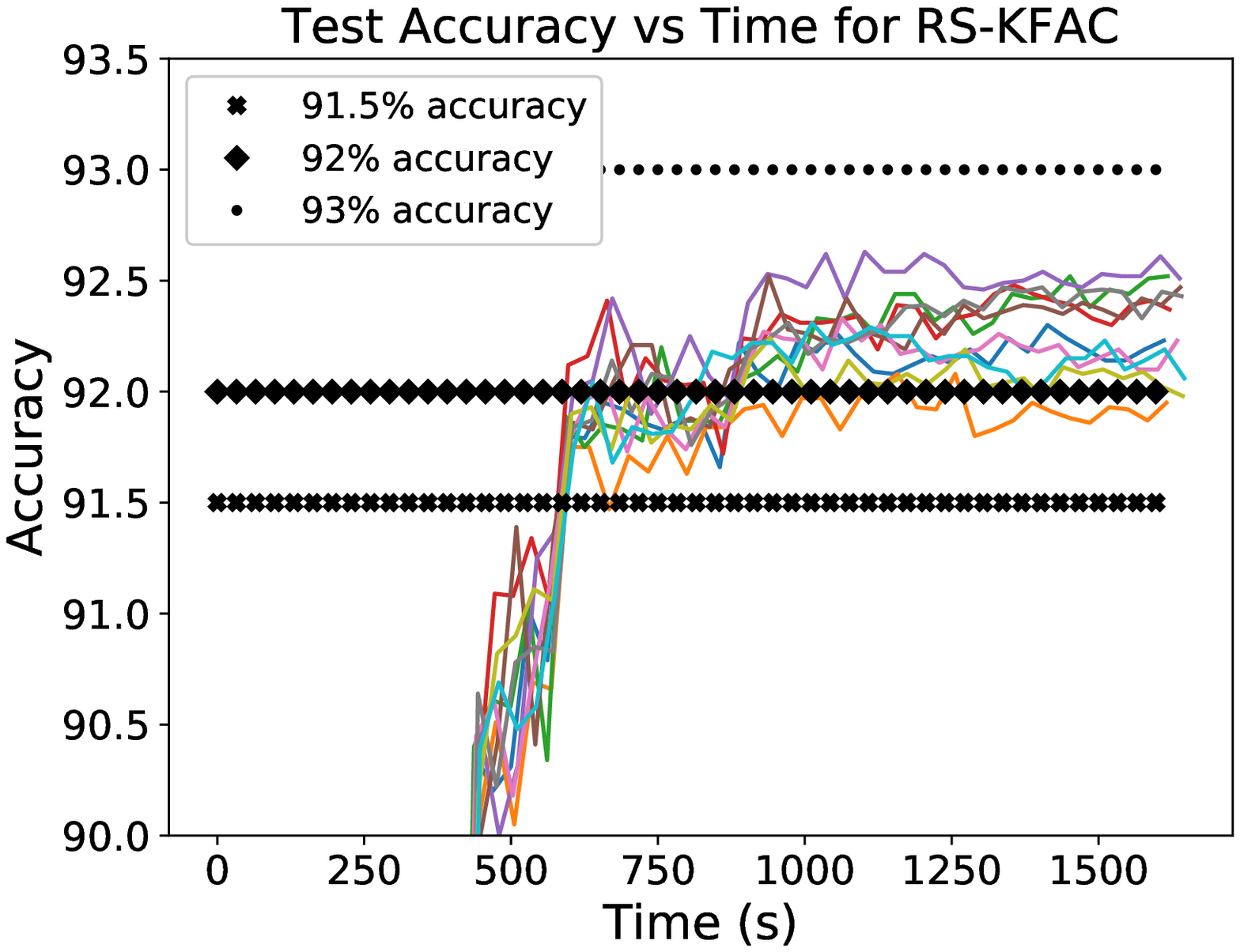}
	\includegraphics[trim={0.1cm 0.15cm 1.6cm 0.9cm},clip,width=0.2935\textwidth]{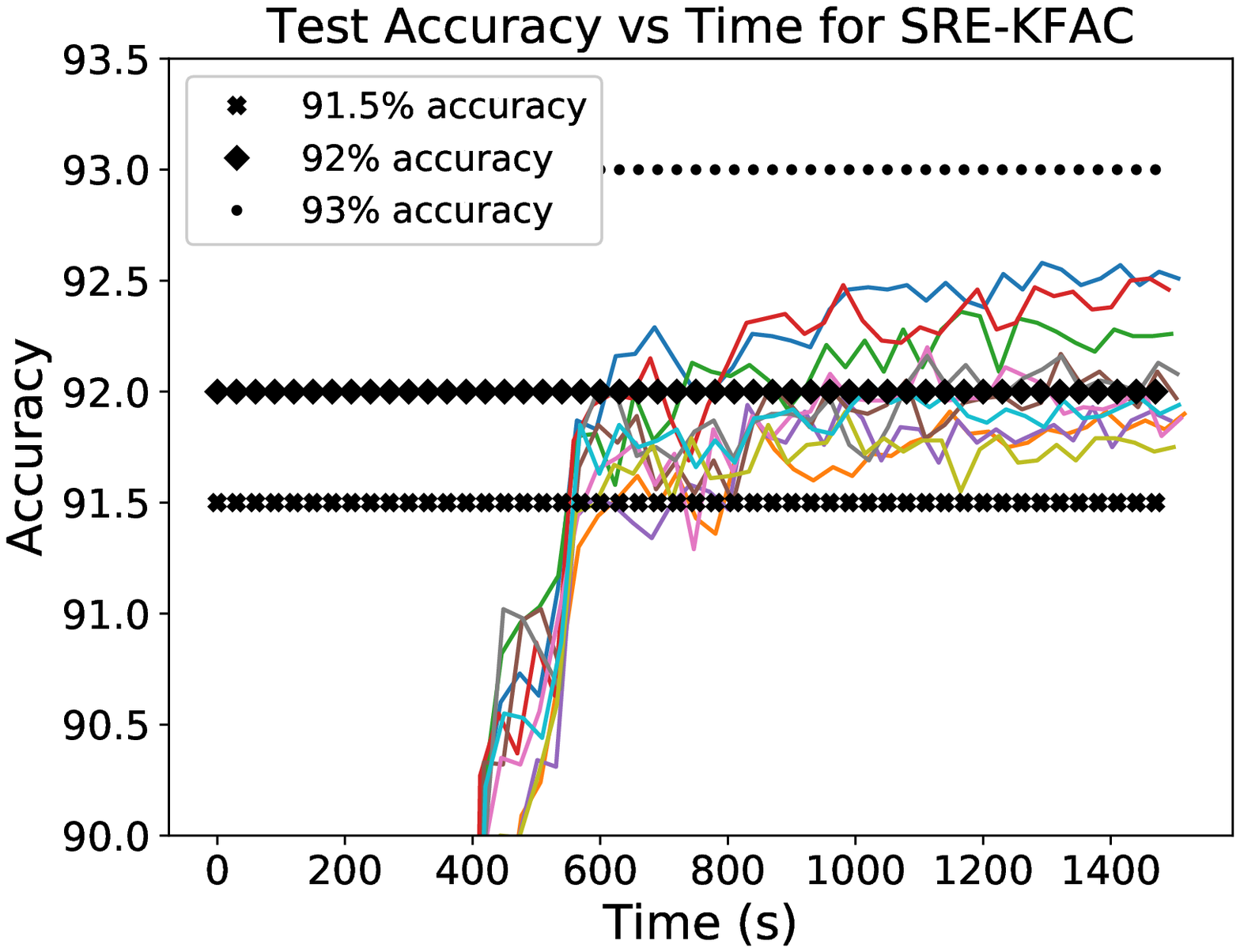}

	\includegraphics[trim={0.1cm 0.15cm 1.6cm 0.9cm},clip,width=0.2935\textwidth]{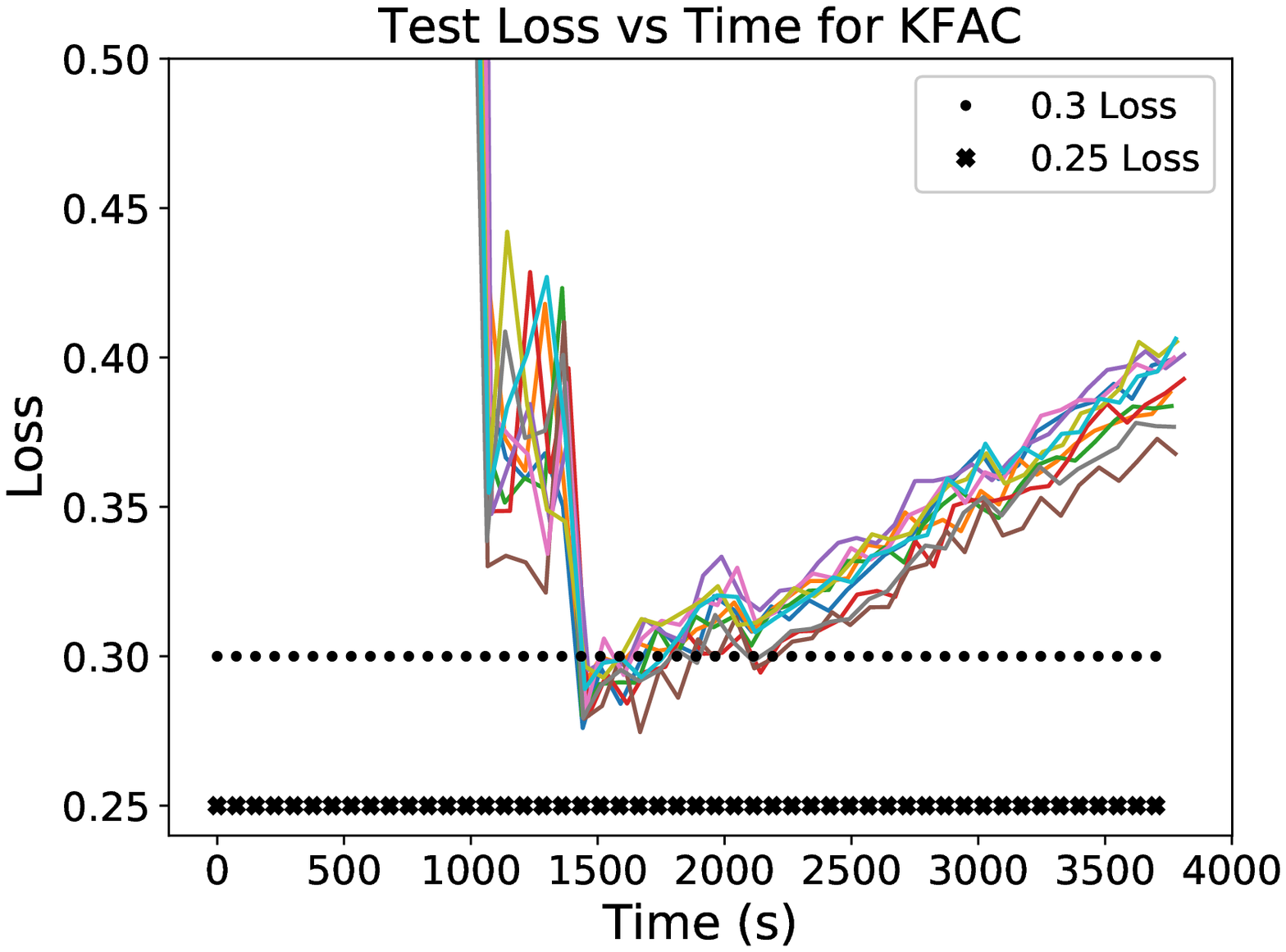}
	\includegraphics[trim={0.1cm 0.15cm 1.6cm 0.9cm},clip,width=0.2935\textwidth]{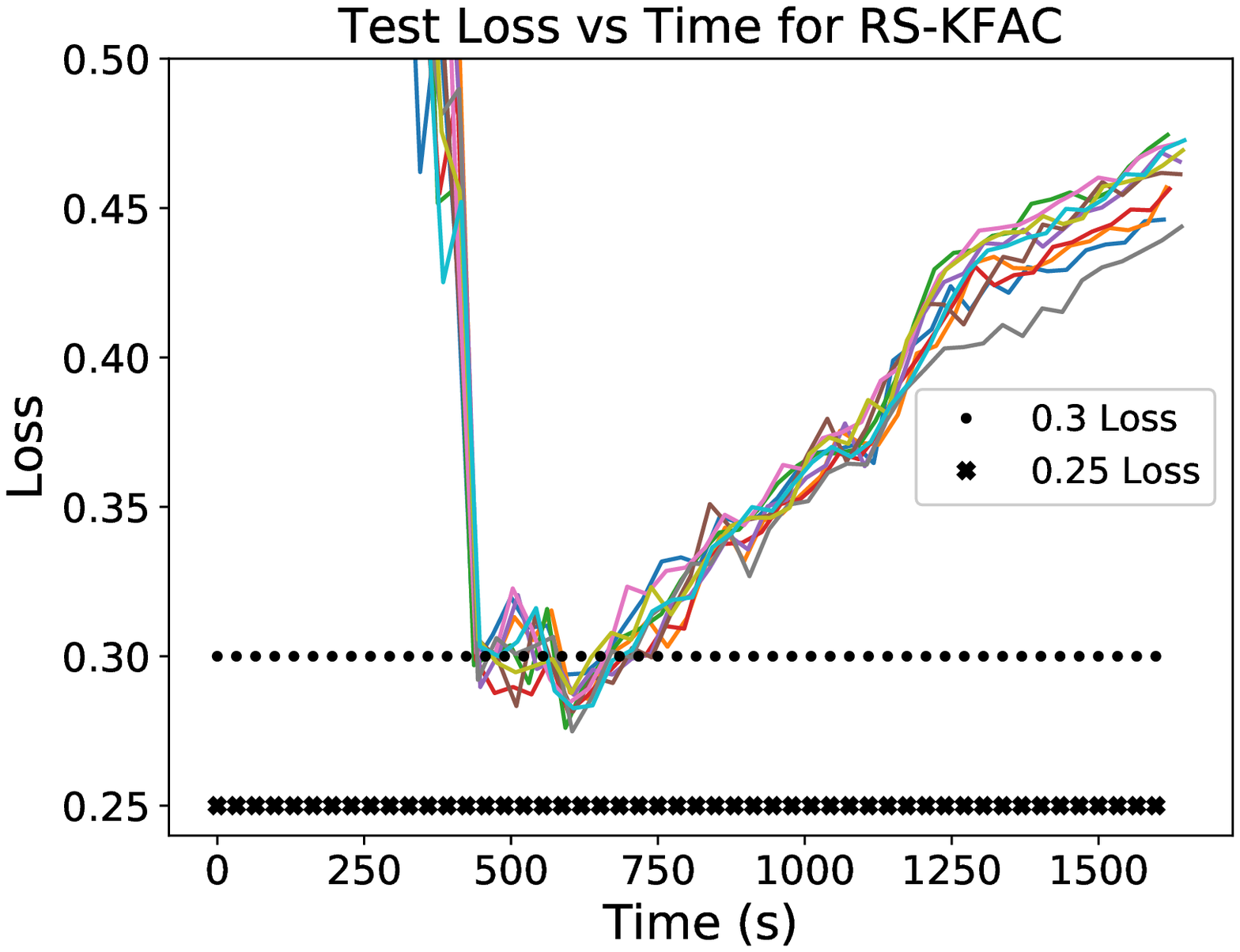}
	\includegraphics[trim={0.1cm 0.15cm 1.6cm 0.9cm},clip,width=0.2935\textwidth]{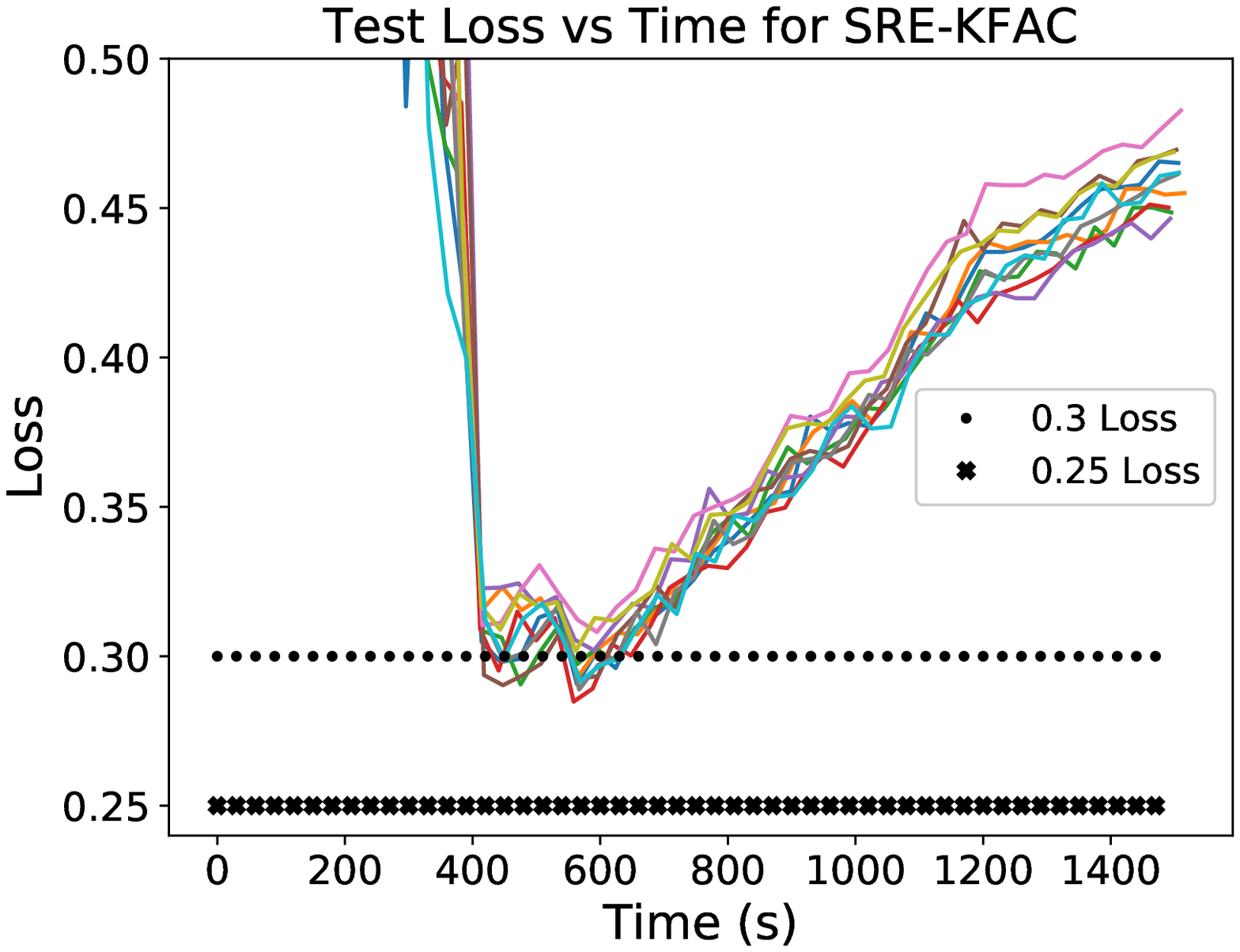}
	
	\includegraphics[trim={0.1cm 0.1cm 1.65cm 0.95cm},clip,width=0.2935\textwidth]{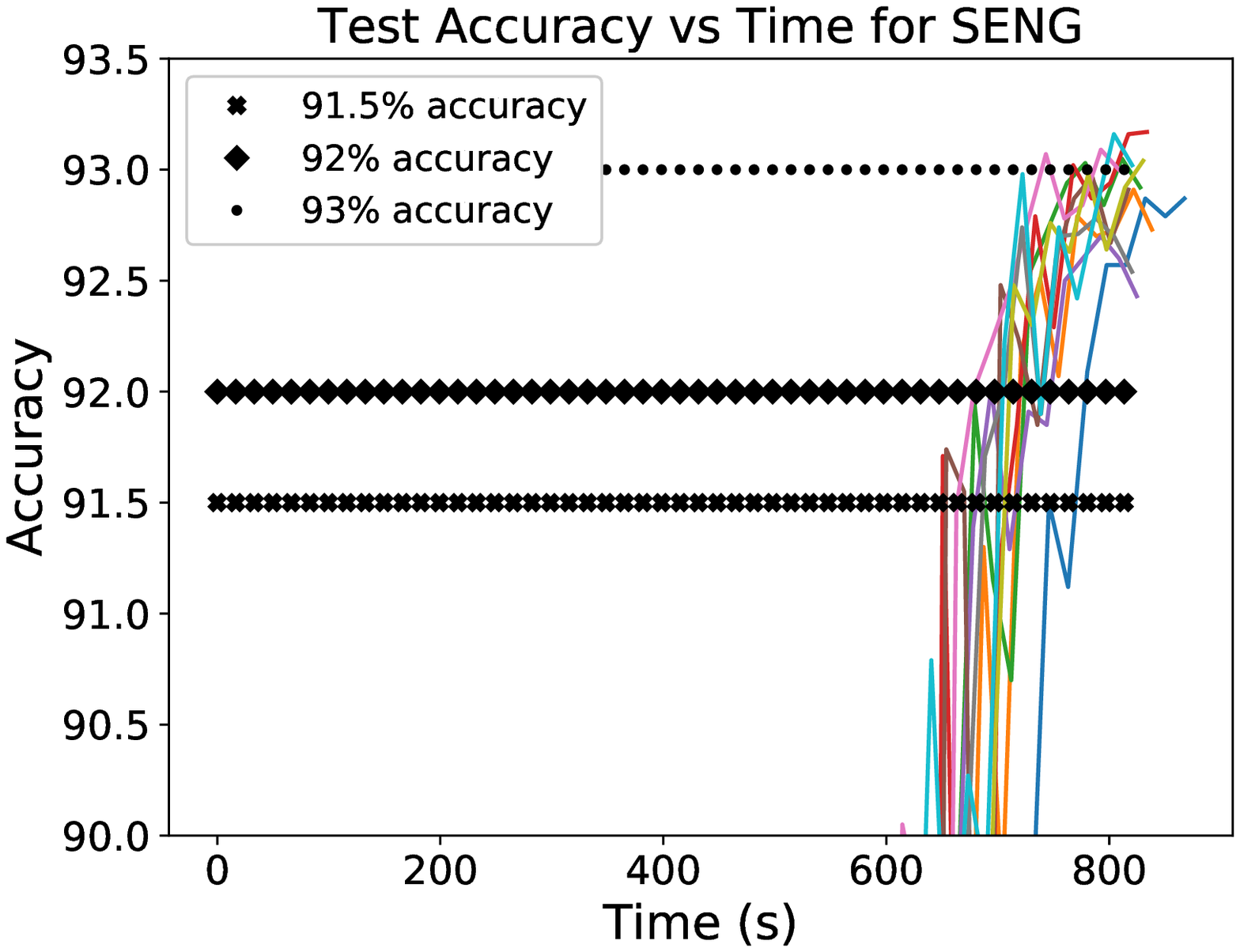}
	\includegraphics[trim={0.1cm 0.1cm 1.65cm 0.95cm},clip,width=0.2935\textwidth]{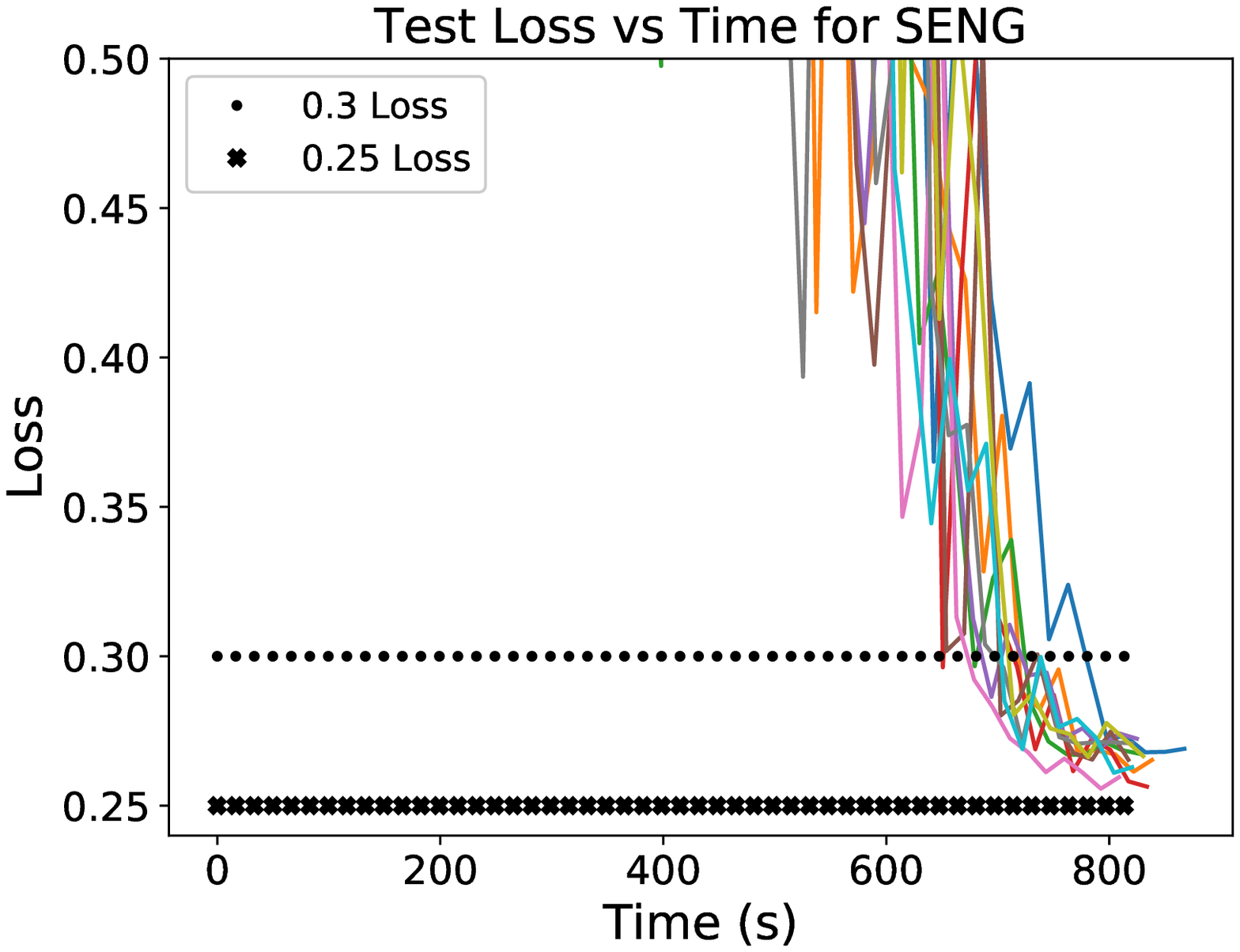}
	
	\caption{CIFAR10 with VGG16\_bn test loss and test accuracy results.} 
	\label{CIFAR10_results_joint_acc_Loss_2}
\end{figure}

\section{Conclusion}
We theoretically observed that the eigen-spectrum of the K-Factors must decay, owing to the associated EA construction paradigm. We then looked at numerical results on CIFAR10 and saw that the decay was much more rapid than predicted by our theoretical worst-case analysis. We then noted that the small eigenvalues are ``washed away'' by the standard K-Factor regularization. This led to the idea that, with minimal accuracy loss, we may replace the full eigendecomposition performed by \textsc{k-fac} with rNLA algorithms which only approximate the strongest few modes. We implicitly answer the question: how many modes?

Importantly, the eigen-spectrum decay was shown (theoretically and numerically) to be such that we only really need to keep a constant number of modes when maintaining a fixed, very good accuracy, irrespectively of what the layer width is! This allowed us to reduce the time complexity from $\mathcal O(d_M^3)$ for \textsc{k-fac} down to $\mathcal O(d_M^2(r+r_l))$ for \textit{Randomized K-FACs}, where $r$ and $r_l$ are constant w.r.t.\ $d_M$ for a fixed desired spectrum cut-off tolerance (for a generic K-factor with layer width $d_M$). We have seen that this complexity reduction from $\mathcal O(d_M^3)$ to $\mathcal O(d_M^2)$ partly closes the gap between \textsc{k-fac} and \textsc{seng} (which scales like $\mathcal O(d_M)$).

 We discussed theoretically that \textsc{rsvd} is more expensive but also more accurate than \textsc{srevd}, and the numerical performance of the corresponding optimizers confirmed this.  Numerical results show we speed up \textsc{k-fac} by a factor of $2.3\times$ in terms of time per epoch, and even had a gain in per-epoch performance. Consequently, target test accuracies were reached about $3.3\times$ faster in terms of wall time. Our proposed \textsc{k-fac} speedups also outperformed the state of art \textsc{seng} (on a problem where it is much faster than \textsc{k-fac}; \cite{SENG}) for $91\%$ and $91.5\%$ target test accuracy in terms of both epochs and wall time. For $92.0\%$ our proposed algorithms only mildly underperformed \textsc{seng}. We argued this could be resolved.

\textbf{Future work:} developing probabilistic theory about eigenspectrum decay which better reconciles numerical results, refining the \textsc{rs-kfac} and \textsc{sre-kfac} algorithms, and layer-specific adaptive selection mechanism for target rank.
\vspace{-2.5ex}

\subsubsection{Acknowledgments}

Thanks to \textit{Jaroslav Fowkes} and \textit{Yuji Nakatsukasa} for useful discussions. I am funded by the EPSRC CDT in InFoMM (EP/L015803/1) together with Numerical Algorithms Group and St.\ Anne's College (Oxford).

\end{document}